\documentclass{article}

\usepackage{arxiv}
\usepackage{times}
\usepackage{graphicx}
\usepackage[figuresleft]{rotating}
\usepackage{svg}
\usepackage{float}
\usepackage{booktabs}
\usepackage{multirow}
\usepackage{tabularx}
\usepackage{longtable}
\usepackage{array}
\usepackage{amsmath,amssymb}
\usepackage{url}
\usepackage{hyperref}
\usepackage{xcolor}
\usepackage{enumitem}
\newlist{tabitemize}{itemize}{1}
\setlist[tabitemize]{
  label=\textbullet,
  leftmargin=*,
  nosep,
  topsep=0pt,
  partopsep=0pt,
  parsep=0pt,
  itemsep=0pt,
  before=\vspace{-0.65\baselineskip},
  after=\vspace{0pt}
}
\newcommand{\agenttypecell}[1]{%
  \parbox[c]{\linewidth}{\centering\textbf{#1}}%
}
\usepackage{tikz}
\usetikzlibrary{arrows.meta,positioning,fit,calc}
\usepackage{soul}
\usepackage{color}
\soulregister\cite7
\soulregister\citep7
\soulregister\citet7
\soulregister\ref7
\soulregister\pageref7

\usepackage[many]{tcolorbox}
\usepackage{lipsum}

\definecolor{bidentitlebg}{RGB}{158,59,255}

\newtcolorbox{conceptbox}[2][]{%
  enhanced,
  skin=enhancedlast jigsaw,
  attach boxed title to top left={xshift=-4mm, yshift=-0.5mm},
  fonttitle=\bfseries\sffamily,
  colbacktitle=blue!45,
  colframe=red!50!black,
  interior style={
    top color=blue!10,
    bottom color=red!10
  },
  boxed title style={
    empty,
    arc=0pt,
    outer arc=0pt,
    boxrule=0pt
  },
  underlay boxed title={%
    \fill[blue!45!white]
      (title.north west) --
      (title.north east) --
      +(\tcboxedtitleheight-1mm,-\tcboxedtitleheight+1mm) --
      ([xshift=4mm,yshift=0.5mm]frame.north east) --
      +(0mm,-1mm) --
      (title.south west) -- cycle;
    \fill[blue!45!white!50!black]
      ([yshift=-0.5mm]frame.north west) --
      +(-0.4,0) --
      +(0,-0.3) -- cycle;
    \fill[blue!45!white!50!black]
      ([yshift=-0.5mm]frame.north east) --
      +(0,-0.3) --
      +(0.4,0) -- cycle;
  },
  title={#2},
  #1
}

\newcounter{sblDef}
\renewcommand{\thesblDef}{\arabic{sblDef}}
\title{
ComBodied Agents: a New Paradigm of Human-Centric Agentic AI
}

\author{
\normalfont
\begin{minipage}{0.95\textwidth}
\large
\centering
\textbf{Qianggang~Ding}\textsuperscript{1,2,*,\ensuremath{\dagger}},
\textbf{Xingyao~Wang}\textsuperscript{3,*},
\textbf{Rui~Feng}\textsuperscript{4},
\textbf{Zhibin~Wang}\textsuperscript{5},
\textbf{Feixiang~Yao}\textsuperscript{5},
\textbf{Kelong~Mao}\textsuperscript{6},
\textbf{Hao~Sun}\textsuperscript{7},
\textbf{Zhiyao~Luo}\textsuperscript{8},
\textbf{Jiankai~Tang}\textsuperscript{9},
\textbf{Lei~Li}\textsuperscript{10},
\textbf{Jiadong~Guo}\textsuperscript{11},
\textbf{Minheng~Ni}\textsuperscript{12},
\textbf{Weicong~Lin}\textsuperscript{13},
\textbf{Chenxi~Yang}\textsuperscript{14},
\textbf{Hongxiang~Gao}\textsuperscript{14},
\textbf{Zhenghua~Chen}\textsuperscript{15},
\textbf{Yang~Bai}\textsuperscript{3},
\textbf{Min~Wu}\textsuperscript{3},
\textbf{Jun~Cheng}\textsuperscript{3},
\textbf{Huazhu~Fu}\textsuperscript{3},
\textbf{Dacheng~Tao}\textsuperscript{16},
\textbf{Bang~Liu}\textsuperscript{1,2,\ensuremath{\dagger}}\\[0.6em]
{\small
\textsuperscript{1}Universit\'e de Montr\'eal, Canada;\quad
\textsuperscript{2}Mila -- Quebec Artificial Intelligence Institute, Canada;\quad
\textsuperscript{3}Institute of Advanced Intelligence and Computing (IAIC), A*STAR, Singapore;\quad
\textsuperscript{4}Nanjing Medical University, China;\quad
\textsuperscript{5}Nanjing University, China;\quad
\textsuperscript{6}Renmin University of China, China;\quad
\textsuperscript{7}University of Cambridge, United Kingdom;\quad
\textsuperscript{8}University of Oxford, United Kingdom;\quad
\textsuperscript{9}Tsinghua University, China;\quad
\textsuperscript{10}National University of Singapore, Singapore;\quad
\textsuperscript{11}The Hong Kong University of Science and Technology, Hong Kong SAR, China;\quad
\textsuperscript{12}The Hong Kong Polytechnic University, Hong Kong SAR, China;\quad
\textsuperscript{13}Southern University of Science and Technology, China;\quad
\textsuperscript{14}Southeast University, China;\quad
\textsuperscript{15}University of Glasgow, United Kingdom;\quad
\textsuperscript{16}Nanyang Technological University, Singapore\\[0.3em]
\textsuperscript{*}\textit{These authors contributed equally to this work.}\\[0.3em]
\textsuperscript{\ensuremath{\dagger}}\textit{Corresponding authors:}
Qianggang Ding
(\href{mailto:qianggang.ding@umontreal.ca}{qianggang.ding@umontreal.ca})
and Bang Liu
(\href{mailto:bang.liu@umontreal.ca}{bang.liu@umontreal.ca})
}
\end{minipage}
}

\date{}

\begin{document}

\maketitle

\begin{abstract}
After an older adult misses a medication dose, a software agent can issue another reminder and an embodied agent can bring the medication. Yet neither action explains whether the person forgot, is confused, is experiencing side effects, or has deliberately refused the medication, nor does it determine what support would be appropriate. This ambiguity exposes a structural gap in Agentic AI. Digital Agents are primarily organized around transformations of software states, while Embodied Agents are organized around transformations of physical states. Neither paradigm makes the evolving state and agency of a person the primary object of modeling, intervention, and evaluation. We introduce Combodied Agents as a human-centered paradigm of Agentic AI that perceives, models, predicts, and supports individual human-state trajectories over time. Software tools, sensors, wearables, robots, and human services serve as action channels rather than final objectives. Although personal assistants, health agents, AI companions, and adaptive human--AI systems instantiate parts of this vision, their capabilities remain fragmented across memory, personalization, sensing, companionship, and domain-specific support. We synthesize these capabilities into a unified closed loop. Event-based multimodal perception reconstructs evidence about meaningful personal events; longitudinal and correctable memory provides temporal context; Personal World Models transform longitudinal event evidence and current context into calibrated distributions over future personal states, observable events, and outcomes under alternative decisions and interventions; and an admissible intervention policy selects proportionate support under constraints of consent, uncertainty, safety, reversibility, and user control. Feedback from the person and environment updates subsequent perception, memory, prediction, and intervention. Rather than requiring an exhaustive Human Digital Twin, this framework uses purpose-bounded, uncertainty-aware, and user-correctable representations of the person. We further organize the design space through human-state targets, relational contexts, and agent roles; examine the transition toward edge-native personal models; and propose scenario-centered evaluation, agency-preservation metrics, benchmark requirements, and governance directions. By shifting Agentic AI from external task completion toward sustained human benefit, Combodied Agents aim to improve health, learning, judgment, capability, relationships, and goal pursuit without treating engagement, dependence, or maximum automation as measures of success.
\end{abstract}

\section{Introduction}

Artificial intelligence (AI) has historically been developed and evaluated primarily for task solving. Earlier task-oriented systems interpreted requests, tracked bounded states, queried domain-specific backends, and returned answers or actions \cite{zhang2020taskoriented}. Large language models (LLMs) and multimodal foundation models have expanded this pattern into general-purpose agents that reason over context, decompose goals, call tools, maintain memory, execute multi-step plans, and adapt from feedback \cite{yao2023react,schick2023toolformer}. The organizing question has shifted from whether AI can answer to how much of an extended task it can complete on its own.

This trend is visible across the two dominant branches of Agentic AI. Digital Agents navigate interfaces, modify code, invoke application programming interfaces (APIs), and coordinate digital workflows \cite{zhou2024webarena,xie2024osworld,jimenez2024swebench}. Embodied Agents connect language and multimodal perception to navigation, manipulation, and physical control \cite{ahn2022saycan,driess2023palme,zitkovich2023rt2}. In both branches, progress means longer task horizons, more reliable action, and less human supervision \cite{liu2024agentbench,wang2024surveyagents}. These advances reduce costs and automate difficult work, but they also normalize an incomplete account of progress: the more work that disappears from the human side, the more capable the agent appears \cite{liu2025human_centered_agents,thinkingmachines2026human}.

What this account misses is that task performance and human development can diverge. AI may improve a document while leaving its author with less understanding, or accelerate a decision while weakening the user's ability to judge its basis. Studies of AI-assisted knowledge work and decision making likewise report reduced cognitive effort and overreliance when people cannot determine when to accept, verify, or reject model outputs \cite{lee2025criticalthinking,vasconcelos2023overreliance}. Task-success metrics capture immediate output, but not the capability or judgment that repeated use leaves with the person.

The concern also has a societal dimension. Human knowledge is often local, tacit, and renewed through participation. Substituting standardized model outputs for those practices may weaken expertise, variety, and resilience. Values create a parallel problem: general-purpose models are developed and aligned in relatively few institutions, whereas users differ in culture, commitments, relationships, and visions of a good life. Concentrating intelligence while weakening people's productive and epistemic agency can therefore concentrate the authority to define both goals and acceptable values \cite{thinkingmachines2026human}.

Personal agents make these tensions especially visible. Memory-enabled assistants, AI companions, health agents, and behavioral coaches increasingly participate in identity, emotion, habits, and care. They can provide continuity and meaningful support, but sustained interaction may also foster dependence, manipulation, social displacement, or inappropriate influence, especially among minors and vulnerable users \cite{liu2025companionship,phang2025affective,fang2025psychosocial,starke2024syntheticrelationships,CommonSense2025Companions}. Automation itself is not the problem; the problem is making substitution the default regardless of its long-term effects on understanding, autonomy, relationships, values, and future capability.

We therefore argue for building \textit{AI for humans}: AI that extends human will and judgment and makes people more capable over time. Human-centered AI has long held that intelligent systems should expand what people can safely and meaningfully do while remaining reliable, understandable, and controllable \cite{shneiderman2020hcai,amershi2019guidelines}. Recent work on Human-Centered Agents makes this principle operational by treating what users retain and develop---understanding, competence, agency, identity, and calibrated reliance---as an outcome alongside task completion \cite{liu2025human_centered_agents}. The optimization target thus expands from isolated task success to immediate benefit plus longitudinal human gain.

AI for humans does not require people in every low-level loop or resist automation for its own sake. Human participation instead becomes a technical design problem: an agent should learn when to act independently, request oversight, or return knowledge and capability to the person \cite{thinkingmachines2026human}. Routine, low-risk, reversible tasks may justify near-complete delegation; learning, health, emotional support, and high-stakes decisions may require explanation, scaffolding, consent, or human escalation. The division of labor should fit the person's goals, state, competence, relationships, and risk while preserving the ability to choose, refuse, correct, and recover.

This commitment motivates \textit{Combodied Agents}: agents organized around beneficial trajectories of human state and agency. Digital tools, wearables, robots, and human services may all serve as action channels, but success is determined by what the person can understand, decide, do, and sustain over time. Section~\ref{sec:action-substrates} develops the distinction from Digital and Embodied Agents.

Combodied Agents share with Human Digital Twins (HDTs) an individual-centered and longitudinal perspective, but the two paradigms define different system boundaries. An HDT aims to maintain a dynamically updated digital representation of a human, or of a use-case-specific aspect of a human, through ongoing data synchronization and feedback. Such a representation may integrate multimodal and multi-scale data to support monitoring, simulation, prediction, and optimization \cite{Lin2024HumanDigitalTwin,LauerSchmaltz2024HDTDesign,Katsoulakis2024DigitalTwinsHealth}.

A holistic, high-fidelity twin of a complete person remains an aspiration rather than a currently attainable system. Human physiology, cognition, behavior, relationships, and context evolve across interacting time scales, while sensing remains incomplete and model validation, synchronization, computation, privacy, and governance remain substantial challenges \cite{LauerSchmaltz2024HDTDesign,Venkatesh2022HealthDigitalTwins,Riahi2025DigitalTwinsDecision}. Combodied Agents therefore do not require an exhaustive digital replica of the person. They maintain purpose-bounded, uncertainty-aware, and user-correctable representations of the aspects needed for an agreed support context, and connect those representations to longitudinal memory, goal negotiation, intervention policies, and feedback. Their primary objective is not maximal representational fidelity, but safe and beneficial participation across the person's evolving contexts while preserving human agency. A domain-specific HDT may supply evidence or predictions to a Combodied Agent, but it does not by itself define the agent's memory, authority, interaction policy, or relationship with the person.

Personal assistants, health agents, AI companions, behavioral coaches, and adaptive human--AI systems already instantiate parts of this vision. What remains missing is a unified paradigm organized around longitudinal human-state trajectories, intervention responses, an adaptive division of labor, and evaluation of both benefit and preserved agency. Combodied Agents translate AI for humans into this technical research program.

Table~\ref{tab:why-combodied-agents} makes this need concrete. Existing agent categories contribute important but partial capabilities: task-oriented agents provide execution, embodied agents provide sensing and actuation, memory and personalized agents provide continuity, and health, learning, companion, and care agents provide domain-specific support. However, no category integrates these capabilities into a longitudinal loop whose primary state is the person and whose success criterion includes capability, autonomy, wellbeing, relationship safety, and agency preservation.

\begin{longtable}{@{}>{\centering\arraybackslash}m{0.14\textwidth}
                    >{\vspace{0pt}\raggedright\arraybackslash}p{0.23\textwidth}
                    >{\vspace{0pt}\raggedright\arraybackslash}p{0.25\textwidth}
                    >{\vspace{0pt}\raggedright\arraybackslash}p{0.28\textwidth}@{}}
\caption[Related agent landscape]{Consolidated related-agent landscape: representative capabilities, limitations, and the human-centered longitudinal support added by Combodied Agents.}
\label{tab:why-combodied-agents}\\
\toprule
\multicolumn{1}{@{}>{\centering\arraybackslash}p{0.14\textwidth}}{\textbf{Agent Type}} &
\multicolumn{1}{>{\centering\arraybackslash}p{0.23\textwidth}}{\textbf{Capabilities}} &
\multicolumn{1}{>{\centering\arraybackslash}p{0.25\textwidth}}{\textbf{Limitations}} &
\multicolumn{1}{>{\centering\arraybackslash}p{0.28\textwidth}@{}}{\textbf{What Combodied Agents Add}} \\
\midrule
\endfirsthead

\caption[]{Consolidated related-agent landscape: representative capabilities, limitations, and the human-centered longitudinal support added by Combodied Agents.}\\
\toprule
\multicolumn{1}{@{}>{\centering\arraybackslash}p{0.14\textwidth}}{\textbf{Agent Type}} &
\multicolumn{1}{>{\centering\arraybackslash}p{0.23\textwidth}}{\textbf{Capabilities}} &
\multicolumn{1}{>{\centering\arraybackslash}p{0.25\textwidth}}{\textbf{Limitations}} &
\multicolumn{1}{>{\centering\arraybackslash}p{0.28\textwidth}@{}}{\textbf{What Combodied Agents Add}} \\
\midrule
\endhead

\midrule
\multicolumn{4}{r}{\emph{Continued on next page}}\\
\endfoot

\bottomrule
\endlastfoot

\agenttypecell{Dialogue Agents}
&
\begin{tabitemize}
\item Intent recognition
\item Dialogue-state tracking
\item Backend querying
\item Bounded task completion
\end{tabitemize}
&
\begin{tabitemize}
\item Episodic rather than longitudinal
\item Restricted to narrow domains
\item Evaluated mainly by task success \cite{zhang2020taskoriented}
\item Limited modeling of evolving user states
\end{tabitemize}
&
\begin{tabitemize}
\item Longitudinal support loop
\item Memory and goal updates
\item Boundary-aware future intervention
\end{tabitemize} \\[5.6\baselineskip]

\agenttypecell{Tool-Use Agents}
&
\begin{tabitemize}
\item Goal decomposition
\item Contextual reasoning
\item Tool and API use
\item Feedback-based planning
\end{tabitemize}
&
\begin{tabitemize}
\item Prioritize task completion over user change
\item Lack intervention-response tracking
\item Weak modeling of user agency
\item Limited safety reasoning beyond tool use \cite{yao2023react,schick2023toolformer,wang2024surveyagents}
\end{tabitemize}
&
\begin{tabitemize}
\item Tool use as support channel
\item Human-state trajectory tracking
\item Outcome, autonomy, and safety focus
\end{tabitemize} \\[5.6\baselineskip]

\agenttypecell{Computer-Use Agents}
&
\begin{tabitemize}
\item Screen understanding
\item Interface operation
\item Form filling
\item Workflow execution
\end{tabitemize}
&
\begin{tabitemize}
\item May automate without human-state awareness
\item Insensitive to fatigue, stress, or overload
\item Weak alignment with user values and obligations
\item Limited assessment of agency loss \cite{zhou2024webarena,xie2024osworld,openai2025cua,anthropic2024computeruse}
\end{tabitemize}
&
\begin{tabitemize}
\item Personal-context-aware execution
\item Calibrated automation and confirmation
\item Agency and safety checks
\end{tabitemize} \\[5.6\baselineskip]

\agenttypecell{Workflow Agents}
&
\begin{tabitemize}
\item Code editing
\item Data analysis
\item File manipulation
\item Process automation
\item Enterprise coordination
\end{tabitemize}
&
\begin{tabitemize}
\item May optimize efficiency at the user's expense
\item Limited attention to cognitive load
\item Weak support for responsibility boundaries
\item Insufficient safeguards for autonomy and control \cite{jimenez2024swebench,drouin2024workarena,hong2025datainterpreter}
\end{tabitemize}
&
\begin{tabitemize}
\item Person-aware planning
\item Cognitive-load and boundary checks
\item Reversible, privacy-preserving control
\end{tabitemize} \\[5.6\baselineskip]

\agenttypecell{Embodied Agents}
&
\begin{tabitemize}
\item Physical perception
\item Navigation
\item Object manipulation
\item Robotic control
\item Safety-aware action
\end{tabitemize}
&
\begin{tabitemize}
\item Limited access to internal human states
\item Weak modeling of life history and relationships
\item Insufficient awareness of vulnerabilities
\item Limited support for longitudinal wellbeing \cite{driess2023palme,zitkovich2023rt2,li2024behavior1k}
\end{tabitemize}
&
\begin{tabitemize}
\item Human-agency support target
\item Person-centered use of sensors
\item Safe, context-aware intervention
\end{tabitemize} \\[5.6\baselineskip]

\agenttypecell{Memory Assistants}
&
\begin{tabitemize}
\item User-fact storage
\item Preference memory
\item Conversation recall
\item Cross-session continuity
\end{tabitemize}
&
\begin{tabitemize}
\item Store facts without modeling trajectories
\item Lack structured intervention history
\item Weak outcome-based memory organization
\item Limited forgetting and correction mechanisms \cite{park2023generative,packer2023memgpt,openai2025memory}
\end{tabitemize}
&
\begin{tabitemize}
\item Longitudinal life memory
\item Events, routines, goals, and relationships
\item Corrections, outcomes, and forgetting
\end{tabitemize} \\[5.6\baselineskip]

\agenttypecell{Personalized Agents}
&
\begin{tabitemize}
\item User profiling
\item Preference adaptation
\item Contextual recommendation
\item Personalized planning
\end{tabitemize}
&
\begin{tabitemize}
\item Often rely on shallow personalization
\item May optimize engagement rather than wellbeing
\item Lack causal models of user change
\item Limited user-contestable adaptation \cite{zhang2025personalization,Xu2026PersonalizedAgents,li2025personalization}
\end{tabitemize}
&
\begin{tabitemize}
\item Personal World Models
\item Action and context response prediction
\item Correctable, safety-aware adaptation
\end{tabitemize} \\[5.6\baselineskip]

\agenttypecell{Companion Agents}
&
\begin{tabitemize}
\item Social presence
\item Emotional continuity
\item Empathic response
\item Rapport building
\item Self-disclosure support
\end{tabitemize}
&
\begin{tabitemize}
\item Risk emotional dependency
\item Blur role and relationship boundaries
\item May reinforce sycophancy or social substitution
\item Lack robust attachment-safety mechanisms \cite{bickmore2005relational,Liu2024ChatbotCompanionship,zhang2024darkside}
\end{tabitemize}
&
\begin{tabitemize}
\item Relationship-safety architecture
\item Scoped memory and role boundaries
\item Dependency detection and escalation
\end{tabitemize} \\[5.6\baselineskip]

\agenttypecell{Health Agents}
&
\begin{tabitemize}
\item Symptom tracking
\item Behavior change
\item Adherence support
\item Emotional coping
\item Chronic-care routines
\end{tabitemize}
&
\begin{tabitemize}
\item Often narrow in health scope
\item Lack causal intervention modeling
\item Weak escalation and uncertainty handling
\item Poor integration with broader life contexts \cite{Merrill2026PHIA,cosentino2024towards,jorke2025gptcoach}
\end{tabitemize}
&
\begin{tabitemize}
\item Multimodal human-state perception
\item Personal baselines and response memory
\item Clinical and personal boundary safeguards
\end{tabitemize} \\[5.6\baselineskip]

\agenttypecell{Learning Agents}
&
\begin{tabitemize}
\item Tutoring
\item Explanation
\item Assessment
\item Learning scaffolding
\item Material adaptation
\end{tabitemize}
&
\begin{tabitemize}
\item May encourage over-reliance
\item Risk weakening independent reasoning
\item Often emphasize short-term performance
\item Limited evaluation of capability growth \cite{wang2025tutorcopilot}
\end{tabitemize}
&
\begin{tabitemize}
\item Capability-growth evaluation
\item Metacognition and self-efficacy support
\item Retention and independent judgment
\end{tabitemize} \\[5.6\baselineskip]

\agenttypecell{Assistive Care Agents}
&
\begin{tabitemize}
\item Reminders
\item Monitoring
\item Companionship
\item Family communication
\item Care coordination
\end{tabitemize}
&
\begin{tabitemize}
\item Risk excessive surveillance
\item Create autonomy--safety tensions
\item Complicate consent and caregiver access
\item Limited support for dignity-preserving intervention \cite{elliq2026features,bellos2025assistive}
\end{tabitemize}
&
\begin{tabitemize}
\item Relation-aware memory
\item Proportional intervention and escalation
\item Contestable dignity-preserving support
\end{tabitemize} \\[5.6\baselineskip]

\agenttypecell{Edge AI Agents}
&
\begin{tabitemize}
\item On-device inference
\item Local sensing
\item Private memory
\item Low-latency personalization
\item Local control
\end{tabitemize}
&
\begin{tabitemize}
\item Privacy gains do not ensure human-centered design
\item Memory and sharing boundaries remain underspecified
\item Lack rich human-state models
\item Weak agency-centered optimization \cite{apple2024intelligence,google2025aicore,chung2026localprivacy,tian2026edgeagents}
\end{tabitemize}
&
\begin{tabitemize}
\item Edge-native personal intelligence
\item User-owned memory and PWMs
\item Local policies with selective cloud use
\end{tabitemize} \\
\end{longtable}

Across the table, three recurring limitations motivate Combodied Agents: existing systems optimize external task or domain outcomes rather than human trajectories; their models of the person are fragmented, shallow, or episodic; and their safety and evaluation criteria rarely measure agency loss, capability growth, relationship effects, or long-term wellbeing. Combodied Agents address these limitations not by simply aggregating more features, but by reorganizing perception, memory, prediction, intervention, and evaluation around the evolving human subject. They add multimodal human-state perception, longitudinal and correctable memory, personal world models (PWMs) of intervention response, calibrated and reversible support policies, and explicit evaluation of human gain and agency preservation.

This paper makes four contributions:

\begin{enumerate}[leftmargin=*,nosep]
    \item We introduce and formally define Combodied Agents as a human-centric paradigm that complements Digital and Embodied Agents by making the evolving human subject and human agency its primary action target.
    \item We develop a closed-loop technical framework connecting multimodal Human State Perception, Longitudinal Memory, PWMs, agency-preserving Intervention Policies, feedback adaptation, and safety constraints.
    \item We synthesize the design space through deployment architecture, human-centered evaluation, and an integrated taxonomy of relationship modes, agent roles, human-state targets, and applications.
    \item We examine the risks, governance requirements, open challenges, and adjacent research fields that shape a responsible research agenda for Combodied Agents.
\end{enumerate}

The remainder of this paper defines Combodied Agents and their closed-loop framework, then examines event-based multimodal perception, PWMs, and cloud-to-edge personal models. It next develops benchmarks, evaluation, taxonomy, applications, risks, governance, and future directions before concluding with the human-agency principle guiding the paradigm.

\section[Foundations of Combodied Agents]{Foundations of Combodied Agents}
\label{sec:combodied-agents}

This section gives the canonical definition of Combodied Agents, distinguishes their action substrate from those of Digital and Embodied Agents, and formalizes the resulting closed loop.

\subsection{Formal Definition and Paradigm Scope}

We formally define the concept of Combodied Agents and their core properties as the following: 

\refstepcounter{sblDef}
\begin{conceptbox}{Definition \thesblDef: Combodied Agents}
\label{def:sbl}

\textbf{Combodied Agents are human-centered intelligent agents that perceive, model, and influence the evolving state of a person through continuous multimodal sensing and longitudinal interaction.} The term \textit{Combodied} combines \textit{Companion} and \textit{Body}, but does not imply a third-party companion whose primary role is conversation or emotional companionship. Instead, a Combodied Agent treats the human body, behavior, cognition, emotion, and surrounding context as its primary perceptual and action domain. By integrating intelligent sensors, longitudinal memory, personalized human-state models, and intervention policies, it continuously observes and understands the individual, anticipates needs or risks, and delivers timely, adaptive, and consent-aware interventions. Its defining characteristic is therefore not merely \textbf{interaction with humans}, but \textbf{perception of humans, understanding of humans, and action on human states}.

\par\medskip
\noindent\makebox[\linewidth][c]{%
    \includegraphics[
        width=0.96\linewidth,
        keepaspectratio
    ]{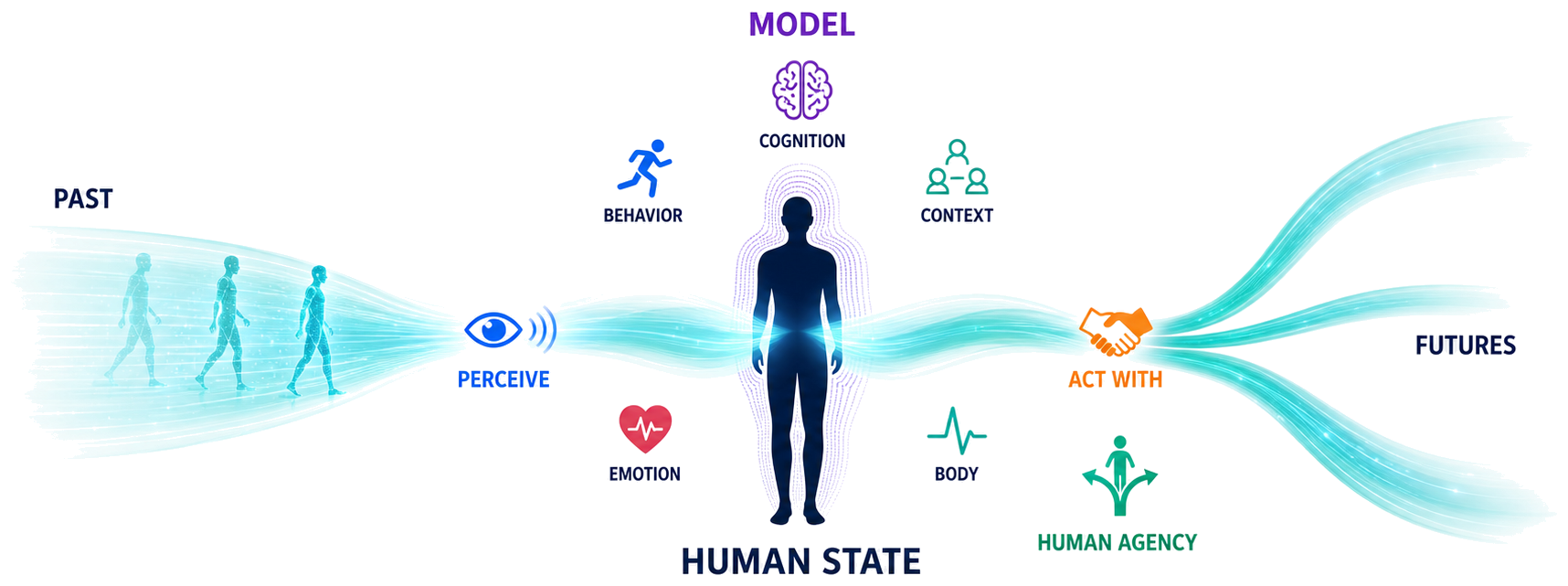}%
}
\par\medskip

The definition establishes five jointly important properties:

\begin{itemize}[leftmargin=*]
    \item \textbf{Human-centric state modeling}: the person, rather than an external task alone, is the primary object of modeling and support.
    \item \textbf{Longitudinality}: the agent reasons across repeated interactions, evolving life contexts, and extended time horizons.
    \item \textbf{Intervention}: the agent may remind, recommend, explain, coach, coordinate, protect, execute, or escalate rather than merely produce responses.
    \item \textbf{Co-agency}: the agent works with the human and adapts the division of labor instead of treating human replacement as the default.
    \item \textbf{Agency preservation}: support must preserve or strengthen autonomy, control, dignity, relationships, and long-term human capability.
\end{itemize}
\end{conceptbox}

These properties define a center of gravity rather than a rigid interface category. A chatbot is not automatically a Combodied Agent because it remembers a name, and a wearable system is not automatically one because it measures a physiological signal. Conversely, a Combodied Agent may act through conversation, software tools, sensors, robots, or human caregivers. What matters is whether these capabilities are integrated around an ongoing, correctable model of the person and used to produce safe, longitudinally beneficial support.

\subsection{Distinguishing Focus: Three Action Substrates}
\label{sec:action-substrates}

\begin{figure}[t]
\centering
\includegraphics[width=\linewidth,height=0.58\textheight,keepaspectratio]{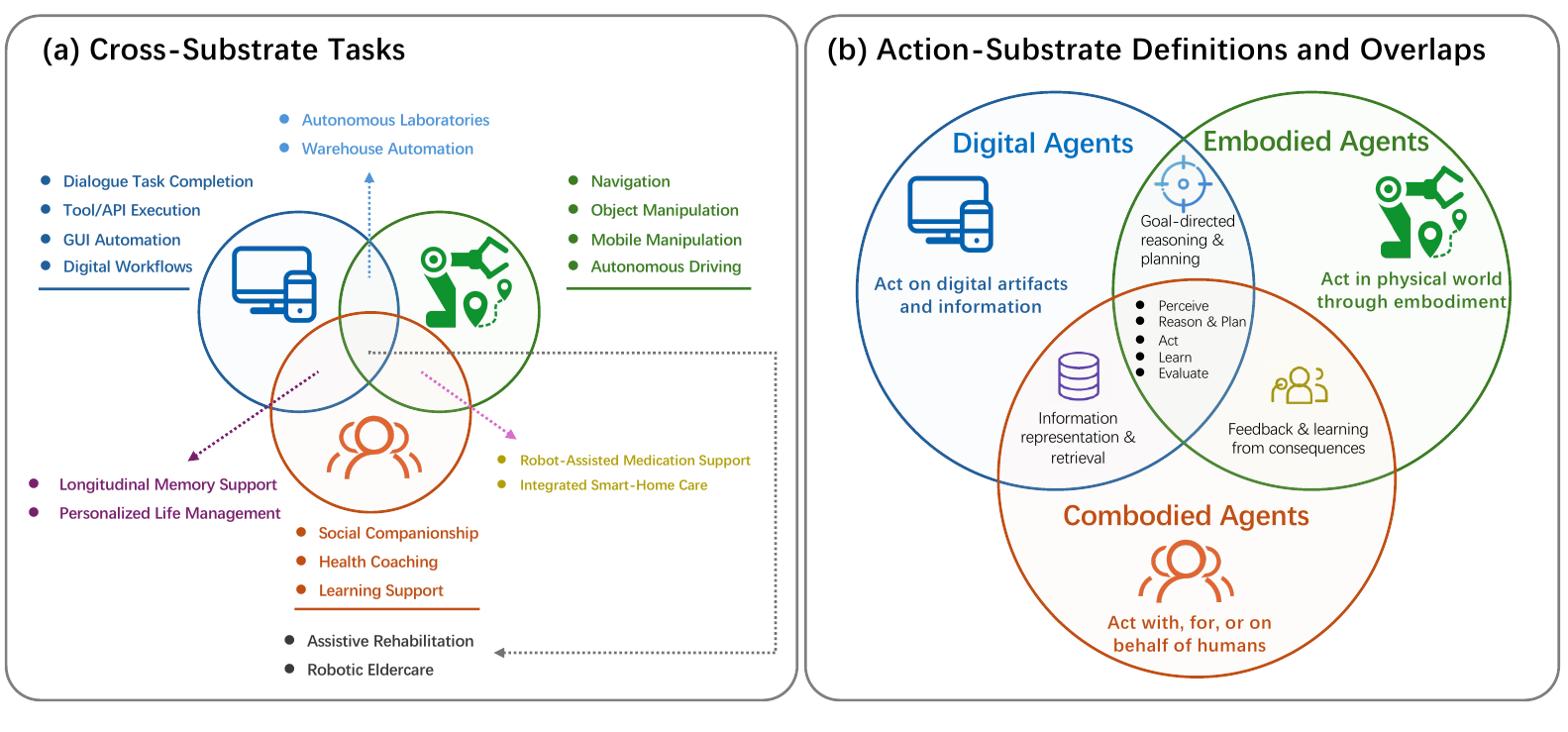}
\caption{Action substrates and representative task configurations in Agentic AI.
\textbf{(a)} Representative substrate-dominant and cross-substrate tasks for Digital, Embodied, and Combodied Agents. The examples are illustrative rather than exhaustive, and their placement indicates which target states and evaluation objectives are jointly involved. \textbf{(b)} The three overlapping centers of gravity are defined by the states that primarily organize modeling, action, and evaluation: digital states and artifacts, physical or simulated physical states, and evolving human states and agency. Pairwise overlaps denote systems that couple digital planning with physical execution, digital support with longitudinal person modeling, or physical assistance with adaptation to human outcomes; the central overlap integrates all three. All three paradigms draw on shared agentic machinery for
perception, representation, reasoning, planning, action, learning, and evaluation.}
\label{fig:three-action-substrates}
\end{figure}

Agentic AI broadly refers to systems that iteratively interpret observations, maintain task-relevant internal states, pursue goals, select and execute actions, and use resulting feedback to guide subsequent behavior. Although particular architectures may not represent goals or plans explicitly, a conceptual agent--environment loop can be expressed as

\begin{equation}
    o_t \rightarrow b_t \rightarrow (g_t, p_t)
    \rightarrow a_t \rightarrow x_{t+1}
    \rightarrow o_{t+1},
\end{equation}

where $o_t$ denotes the agent's observation at time $t$, $b_t$ its internal belief or task-relevant state, $g_t$ its current goal, $p_t$ an optional plan, $a_t$ the selected action, and $x_{t+1}$ the resulting state of the relevant external environment or target domain. The next observation $o_{t+1}$ is generated from this updated state. Feedback may update the agent's internal state, revise its plan, or change its subsequent actions, without necessarily implying continual learning or model-parameter updates.

One useful and complementary way to distinguish agent paradigms is by the class of target states that primarily organizes their modeling, decision making, intervention, and evaluation. We call this class the agent's \textit{action substrate}. The action substrate is not necessarily identical to the agent's interface or immediate actuation channel: an agent may act through software tools, sensors, robots, or communication while ultimately seeking to support a different target-state trajectory. This perspective complements, rather than replaces, taxonomies based on model architecture, autonomy, embodiment, tools, memory, or perceptual modality.

For the purposes of this paper, we distinguish three overlapping and non-exhaustive centers of gravity. Digital Agents are primarily organized around transformations of digital states; Embodied Agents around transformations of physical or simulated physical states; and Combodied Agents around modeling and supporting the evolving human state while preserving or strengthening human agency over time. Figure~\ref{fig:three-action-substrates} summarizes these three centers of gravity.

The overlaps in Figure~\ref{fig:three-action-substrates} represent cross-substrate task configurations rather than uncertainty about category boundaries. Autonomous laboratories, for example, couple digital planning with physical execution; longitudinal memory support combines digital action with person modeling; assistive   habilitation combines physical intervention with adaptation to human outcomes; and robot-assisted medication support may integrate all three substrates. Such intersections are expected because research programs frequently extend beyond their historical centers of gravity by incorporating adjacent capabilities, action channels, and evaluation objectives. A system may operate across multiple platforms or embodiments while still pursuing one primary outcome for the user. Agents should therefore be classified by the target user state and success criteria that guide their behavior, rather than by their interface, underlying model, or mode of action.

Digital Agents operate primarily on digital states. Browser and graphical user interface (GUI) agents navigate interfaces; coding agents modify repositories; tool-use, data-analysis, and workflow agents invoke APIs and transform digital artifacts \cite{deng2023mind2web,zhou2024webarena,xie2024osworld,jimenez2024swebench,qin2024toollm,drouin2024workarena}. Their center of gravity is an externally represented task state, and they are usually evaluated by correctness, completion, efficiency, policy compliance, and robustness in digital environments.

Embodied Agents operate primarily on physical or simulated physical states. Robotic agents, autonomous vehicles, household robots, industrial systems, and navigation agents connect perception, planning, and control to movement and manipulation \cite{driess2023palme,zitkovich2023rt2,hu2023planning,yenamandra2024homerobot,oneill2024openx,puig2023habitat}. Their center of gravity is the physical environment, and evaluation focuses on task success, safety, control reliability, and generalization across bodies and environments.

These categories are indispensable but incomplete for systems whose principal concern is the evolving human subject. A personal agent may use a browser to schedule care or an embodied sensor to detect fatigue, yet neither the webpage nor the sensorimotor state is its ultimate target. The relevant outcome is whether the intervention helps the person regulate health, understand a decision, sustain a relationship, develop a capability, or pursue a valued goal. Combodied Agents name this third center of gravity: human agency and its physiological, cognitive, behavioral, emotional, social, and goal-directed conditions over time. The three paradigms therefore overlap in implementation while differing in what their actions are ultimately for.

\subsection{High-Level Core Capabilities}

\begin{figure}[t]
\centering
\includegraphics[width=\linewidth]{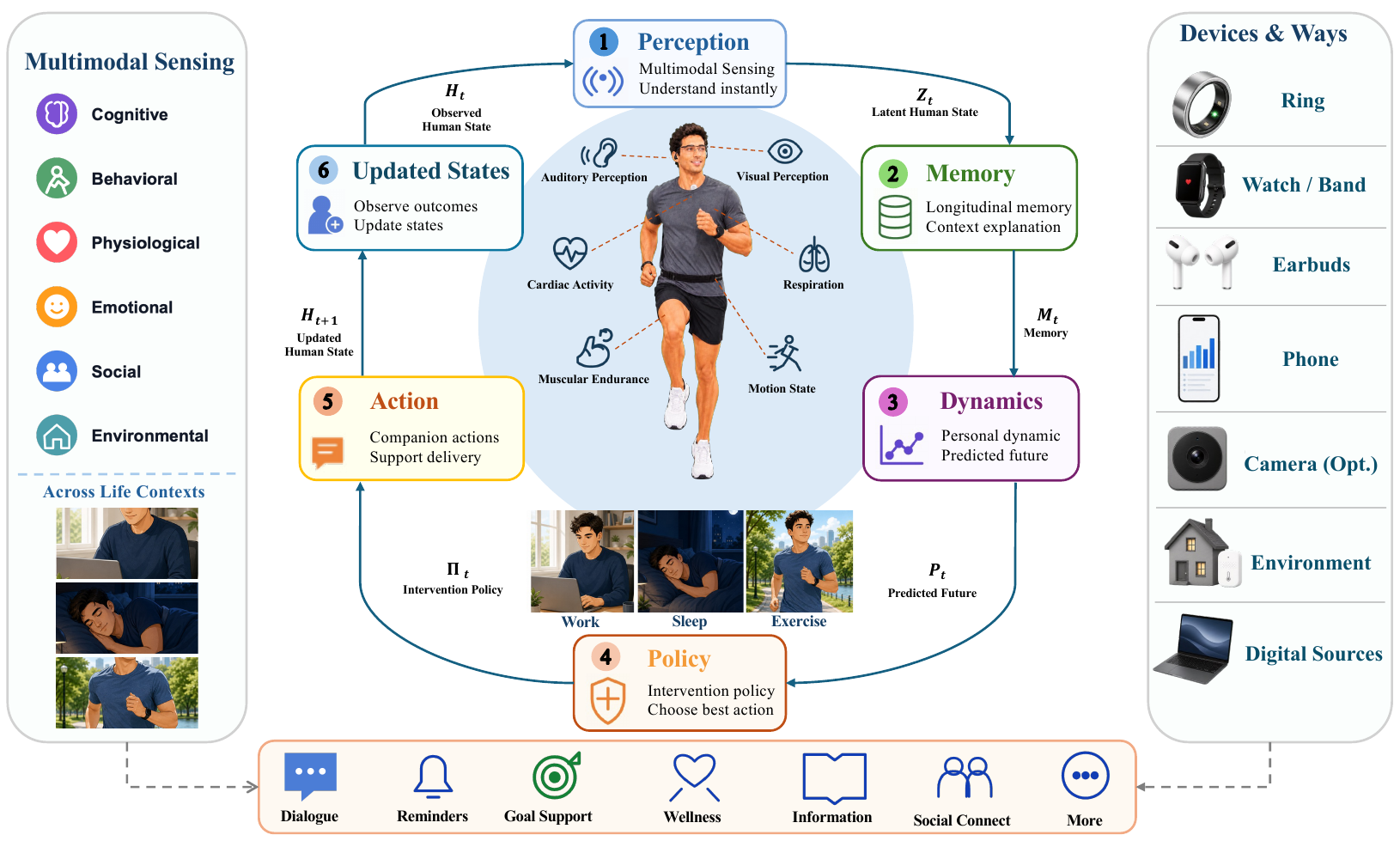}
\caption{Functional organization of a Combodied Agent. Multimodal observations support human-state estimation; longitudinal memory supplies relevant temporal context; a personal world model predicts state--event--outcome trajectories under alternative decisions, interventions, and contextual changes; and an intervention policy selects and delivers appropriate support. Feedback from the person and environment closes the loop. Safety, uncertainty, consent, and user control constrain every stage.}
\label{fig:combodied-agent-overview}
\end{figure}

The definition above implies four connected capabilities, summarized in Figure~\ref{fig:combodied-agent-overview}.

\begin{enumerate}[leftmargin=*]
    \item \textbf{Human-State Perception} estimates current personal state from incomplete, noisy, and multimodal evidence.
    \item \textbf{Longitudinal Memory} organizes events, goals, relationships, interventions, outcomes, and corrections across time.
    \item \textbf{Personal World Modeling} predicts person-specific state--event--outcome trajectories under alternative user actions, agent interventions, and contextual changes.
    \item \textbf{Intervention Planning and Delivery} selects whether, when, and how to provide proportionate support or human escalation.
\end{enumerate}

These are functional components rather than additional defining properties. Human-centricity and longitudinality determine what they model; co-agency and agency preservation constrain the complete loop.

\subsection{Closed-Loop Architecture and Formalization}
\label{sec:combodied-framework}

The following formalization specifies the information flow among human-state inference, longitudinal memory, prediction, intervention, and feedback.

\paragraph*{Latent Human State and Internal Representation.}

Let $H_t$ denote the unobservable human state at time $t$, and let $D_{\leq t}$ contain the event-evidence records reconstructed from observations up to $t$. The agent maintains an uncertainty-bearing posterior representation and retrieves decision-relevant evidence:

\begin{align}
    Z_t
    &\sim
    q_\phi\!\left(\cdot \mid D_{\leq t}, M_t, C_t\right), \label{eq:state-posterior}\\
    R_t
    &=
    \mathrm{Read}
    \left(
        M_t; Z_t, C_t, G_t
    \right).
\end{align}

Here $C_t$ is current context, $M_t$ longitudinal personal memory, $G_t$ the user's current goals, and $R_t$ the retrieved evidence relevant to the decision. Section~\ref{sec:multimodal-perception} defines the event-evidence interface, and Section~\ref{sec:pwm} defines action-conditioned trajectory prediction from this state.

\paragraph*{Prediction and Intervention Selection.}

The PWM predicts multidimensional future outcomes under candidate interventions, and the policy chooses only among actions admissible under consent, stated boundaries, safety constraints, uncertainty thresholds, reversibility, and escalation requirements. Section~\ref{sec:pwm} gives the single canonical prediction and selection formulation; its admissible set includes non-intervention, clarification, confirmation, and referral.

\paragraph*{Human-State Transition and Feedback.}

The subsequent human state is not determined by the agent's action alone. It also depends on the person's own actions, contextual changes, and exogenous influences:

\begin{align}
    H_{t+1}
    &\sim
    T_H
    \left(
        \cdot
        \mid
        H_t,
        a_t^{\mathrm{agent},*},
        a_t^{\mathrm{user}},
        \Xi_t
    \right), \\
    O_{t+1}
    &\sim
    \Omega
    \left(
        \cdot
        \mid
        H_{t+1}, C_{t+1}
    \right), \\
    M_{t+1}
    &=
    \mathrm{Update}
    \left(
        M_t,
        O_{t+1},
        a_t^{\mathrm{agent},*},
        Y_{t+1},
        F_{t+1}
    \right).
\end{align}

Here $T_H$ denotes the human-state transition process, $a_t^{\mathrm{user}}$ the user's own actions, $\Xi_t$ exogenous influences, $\Omega$ the observation process, and $F_{t+1}$ explicit or implicit feedback. The new evidence updates memory and may subsequently revise the state estimator, PWM, or Intervention Policy. This closes the longitudinal loop without assuming that an intervention has a deterministic or immediately observable effect on the person.

\paragraph*{Longitudinal Memory as Temporal Evidence.}

Longitudinal Memory records temporally extended evidence about states, experiences, goals, preferences, relationships, interventions, and outcomes. Each memory item may include content, timestamp, source, modality, confidence, sensitivity, relevant goals, intervention and outcome links, and a retention policy. Table~\ref{tab:longitudinal-memory-components} summarizes the principal functions.

\begin{table}[t]
\centering
\caption{Main components of Longitudinal Memory in Combodied Agents.}
\label{tab:longitudinal-memory-components}
\begin{tabular}{p{0.3\linewidth}p{0.65\linewidth}}
\toprule
\textbf{Memory Component} & \textbf{Function} \\
\midrule
Episodic memory & Records concrete events, interactions, and user experiences. \\
Semantic person memory & Stores relatively stable facts, preferences, values, and boundaries. \\
Trajectory memory & Tracks changes in health, behavior, emotion, cognition, and routines. \\
Goal and commitment memory & Maintains user goals, plans, obligations, and promises. \\
Relationship memory & Represents relevant family, caregiver, clinician, work, or social ties. \\
Intervention-response memory & Links reminders, recommendations, coaching, or escalation to acceptance, rejection, benefit, or harm. \\
User-control memory & Stores restrictions, do-not-remember rules, corrections, and deletion requests. \\
\bottomrule
\end{tabular}
\end{table}

Memory writing, consolidation, retrieval, and correction must preserve provenance and uncertainty and remain inspectable and correctable by the user \cite{zhong2024memorybank,packer2023memgpt,du2026memory}. In particular, intervention-response memory supplies evidence for adapting future support rather than merely repeating a previously generated action.

\paragraph*{Intervention Semantics.}

The policy operates over actions that differ in target, intensity, reversibility, and required authority. Table~\ref{tab:companion-action-space} summarizes this action space; its entries are possible support modes rather than an assumption that intervention is always warranted.

\begin{table*}[t]
\centering
\caption{Combodied Agent action space.}
\label{tab:companion-action-space}
\begin{tabular}{lll}
\toprule
\textbf{Action Type} & \textbf{Primary Target} & \textbf{Examples} \\
\midrule
Inform & cognition & explain, summarize, educate, interpret data \\
Remind & attention and execution & medication reminder, meeting reminder, rest reminder \\
Recommend & decision-making & suggest options, resources, plans, or next steps \\
Coach & capability and habit & train, give feedback, review progress, encourage \\
Nudge & behavioral tendency & gently steer toward healthier or safer choices \\
Reflect & self-understanding & help review emotions, goals, patterns, and trade-offs \\
Coordinate & social and institutional context & contact caregivers, clinicians, colleagues, or services \\
Protect & safety and autonomy & detect scams, warn about risk, block unsafe actions \\
Escalate & high-risk states & refer to human expert, clinician, caregiver, or emergency support \\
Execute & delegated external action & book appointments, send messages, operate software tools \\
\bottomrule
\end{tabular}
\end{table*}

The policy should not always intervene. Depending on uncertainty, consent, and risk, the appropriate action may be to remain silent, request clarification or confirmation, reduce intervention intensity, or escalate to an authorized human expert. Safety and human control therefore constrain the entire loop: perception should avoid unsupported inference; memory should support inspection, correction, and deletion; action selection should enforce proportionality and reversibility; and adaptation should not optimize engagement or dependence at the expense of autonomy and wellbeing \cite{liu2025companionship,starke2024syntheticrelationships}. Deployment implications for keeping these sensitive computations under user control are developed separately in Section~\ref{sec:edge-personal-models}.

\section{Event-Based Multimodal Perception}
\label{sec:multimodal-perception}

Combodied Agents require multimodal perception \cite{Durante2024AgentAI} because human agency is rarely visible in a single utterance, record, or sensor reading. The challenge is not to accumulate as much personal data as possible, but to recover useful fragments from long-term, intermittent, noisy, and unevenly sampled data; identify events that matter; and retain enough evidence to explain why an event is relevant to the user's state, goals, safety, or future support. A late-night message, a missed medication reminder, a short walk, an unusual pause in speech, or a caregiver note may be weak evidence on its own. When placed in personal and temporal context, however, several such fragments may form a meaningful account of change.

We define \textit{multimodal perception} in Combodied Agents as event-based personal data perception: the acquisition, filtering, alignment, and interpretation of fragmentary personal data for human-state understanding and longitudinal modeling. Its output is not an unrestricted stream of raw data. Observations are first checked for quality and segmented into candidate events; only then can they become \textit{event-evidence records} describing what was observed, when and how it was acquired, which interpretation it supports, what uncertainty remains, and whether it is relevant enough to influence memory or intervention.

Acquisition is part of this definition because the same nominal modality can have different meanings depending on how it was obtained. A user may actively describe an experience, respond to a context-triggered prompt, carry a sensing device, wear a sensor in contact with the body, enter an instrumented environment, or authorize access to an institutional record. We refer to these configurations as user-reported, device-mediated, on-body or contact, ambient or contactless, and institutional acquisition. Their sampling patterns range from continuous and periodic sensing to episodic, event-triggered, context-triggered, user-initiated, and query-on-demand collection. These distinctions affect coverage, burden, privacy, and evidential strength, and should remain visible downstream.

Personal data are also longitudinal but discontinuous. A companion may receive only sparse glimpses of a routine or condition, separated by hours, days, or weeks. Perception must therefore identify transitions, deviations, social episodes, safety-relevant events, and responses to previous interventions rather than classify isolated samples. At the same time, inferred states must remain linked to observable fragments. Multimodal evidence can make an interpretation more plausible, but does not turn it into an uncontestable fact about the person.

\subsection{Language and Textual Signals}

Language is the most direct channel through which users can state goals, preferences, commitments, emotions, experiences, and boundaries. Relevant evidence may appear in conversations with the agent, messages, diaries, ecological momentary assessments, questionnaires, notes, task histories, or corrections to an earlier interpretation. Some of these data are deliberately authored for the agent; others are created for a different purpose and become available through an authorized application interface or user upload. Text may also be dictated and transcribed or extracted from documents, but its acquisition history should remain visible: a spontaneous statement, an answer to a prompt, and an automatically imported note do not carry the same evidential meaning.

For longitudinal support, the important textual events are usually changes rather than isolated facts. A user may formulate a new goal, revise a preference, withdraw consent, report an unsuccessful intervention, or correct something that the agent previously stored. Experience-sampling research shows the value of collecting reports close to the moment of experience \cite{shiffman2008ecological}, while systems such as MemoryBank and benchmarks such as LongMemEval investigate continuity across extended interaction histories \cite{zhong2024memorybank,wu2025longmemeval}. A Combodied Agent adds a governance requirement to this line of work: the person must be able to distinguish what they explicitly said from what the system inferred, and later corrections should update the personal model without erasing the provenance of the original claim.

Text is explicit but not necessarily objective. Reports may be incomplete, socially desirable, emotionally amplified, retrospectively distorted, or shaped by the agent's question. Language is therefore particularly valuable for representing user-authorized goals and boundaries, but it remains one source of evidence rather than a complete account of behavior or wellbeing.

\subsection{Speech and Audio Signals}

Text records what is said but removes much of how it is said. Speech restores temporal and paralinguistic information through pitch, intensity, rhythm, pauses, hesitation, articulation, fluency, and voice quality. The surrounding audio may also contain events that are not linguistic, including alarms, coughing, crying, collisions, or changes in the sound environment.

These signals reach an agent through technically and socially different configurations. Near-field microphones in phones, computers, earbuds, and hearing devices mainly capture deliberate interaction. Contact or throat microphones reduce some environmental interference but require the device to be worn. Far-field microphone arrays in rooms, vehicles, or smart speakers provide broader coverage, while also increasing the likelihood of recording bystanders and activities unrelated to the support purpose. Conversation-bound or voice-triggered acquisition is therefore materially different from continuous ambient listening, even when both ultimately produce an audio stream.

Speech research has traditionally separated voice activity detection, speaker diarization, speech recognition, acoustic feature extraction, and environmental sound detection. GeMAPS and eGeMAPS were introduced to make acoustic analysis more reproducible \cite{eyben2016gemaps}, while MERBench places speech alongside linguistic and visual evidence in multimodal emotion recognition \cite{lian2024merbench}. These resources provide useful components, but longitudinal companionship introduces a different question: whether an observed change is unusual for this particular speaker and persists long enough to constitute a relevant event. Repeated hesitation during a familiar task may be informative; one pause in a noisy conversation is usually not.

Acoustic features remain indirect evidence. Pauses may result from reflection, distraction, fatigue, network delay, or cognitive difficulty, and voice characteristics vary with language, health, device, and environment. Local feature extraction, short-lived audio buffers, explicit activation indicators, and reliable speaker attribution are consequently more appropriate than retaining raw recordings by default.

\subsection{Vision-Based Sensing}

Visual sensing broadens perception from the interaction channel to visible behavior and surrounding activity. Device-facing cameras on phones and computers capture expressions, gaze, and posture during direct interaction. Wearable cameras or smart glasses provide a first-person view of activities and object use, while fixed cameras offer a wider view of rooms, movement, and multi-person events. Depth, infrared, thermal, eye-tracking, and event-based sensors extend these arrangements when illumination, distance, temporal resolution, or particular privacy constraints matter.

The unit of interest is rarely an isolated frame. A useful system needs to recognize episodes such as beginning a meal, taking medication, leaving home, completing an exercise, interacting with another person, or changing behavior after an intervention. OpenFace demonstrates real-time extraction of facial landmarks, head pose, facial action units, and eye gaze from conventional cameras \cite{baltrusaitis2016openface}. Ego4D moves toward everyday first-person perception through large-scale egocentric video and tasks involving episodic memory, object interaction, social activity, and anticipation \cite{grauman2022ego4d}. These lines of research suggest components for Combodied Agents, but they do not by themselves solve the longitudinal problem of deciding which visual episodes matter to a particular person.

The limits of visual interpretation are especially important. Reduced facial movement may reflect concentration, culture, disability, fatigue, lighting, or camera position rather than low affect. Occlusion, identity errors, and context-dependent gestures introduce further ambiguity. A visual event should therefore retain the acquisition viewpoint, image quality, subject-attribution confidence, and presence of other people. Where possible, raw images should be processed locally and replaced by purpose-specific event descriptions. The fact that a camera can capture a household scene does not mean that every visible person or object belongs in the user's memory.

\subsection{Physiological and Biochemical Signals}

Physiological sensing appears closer to bodily state than language or vision, but this proximity does not eliminate ambiguity. Watches, rings, chest straps, earbuds, patches, headbands, cuffs, beds, and clinical monitors can provide heart rate, heart-rate variability, electrocardiography, photoplethysmography, blood oxygen saturation, electrodermal activity, respiration, temperature, sleep-related signals, blood pressure, and selected neurological or muscular measurements. Biochemical sensing adds continuous glucose monitors and emerging sensors for sweat, saliva, and interstitial-fluid biomarkers. Some measurements require contact or minimally invasive devices; others, including remote photoplethysmography, thermal sensing, and radar-based respiration estimation, can be obtained without direct contact.

The acquisition configuration changes both burden and reliability. Consumer wearables provide convenient longitudinal coverage but are affected by device placement, motion, skin contact, battery state, firmware, and proprietary preprocessing. Chest straps and clinical equipment may provide higher-fidelity measurements for a narrower period. Skin patches can combine sensing functions: hybrid systems have, for example, demonstrated simultaneous ECG and sweat-lactate measurement \cite{imani2016hybrid}. These differences should not disappear when measurements are converted into a common numerical time series.

For a longitudinal companion, isolated thresholds are generally less useful than personal baselines, recovery trajectories, and persistent changes across days or weeks. Relevant events may include sustained sleep deterioration, an unusual resting heart rate, repeated glucose excursions, delayed recovery following activity, or disagreement between subjective symptoms and sensor measurements. Health-LLM and PHIA illustrate how language-model-based systems can reason over personal health and wearable data \cite{kim2024health,Merrill2026PHIA}. Their results demonstrate a reasoning interface, however, rather than clinical validation of every inferred state.

Motion, medication, hydration, temperature, illness, and sensor contact can all change a physiological reading. The same elevated heart rate may be expected during exercise and concerning at rest. Physiological evidence can support awareness, self-management, and preparation for professional care, but should not be promoted to diagnostic truth without suitable calibration, clinical validation, and escalation boundaries.

\subsection{Motion and Behavior Monitoring}

Motion sensing sits between bodily measurement and situational context. Accelerometers, gyroscopes, magnetometers, and barometers on phones or wearables record movement and orientation; satellite positioning, Bluetooth, Wi-Fi, and ultra-wideband add mobility and proximity information. Homes and care environments may contribute passive infrared sensors, pressure mats, instrumented furniture, smart appliances, cameras, or millimeter-wave radar. These arrangements differ in what they observe: a wrist sensor follows part of the body, a phone is useful only while carried, and an ambient sensor observes activity within a particular space.

Human-activity-recognition research has developed methods for converting such streams into actions and routines. OPPORTUNITY combines wearable, object, and ambient sensing for activity recognition \cite{roggen2010opportunity}; DeepConvLSTM illustrates end-to-end sequence modeling from wearable data \cite{ordonez2016deep}; and recent surveys synthesize multimodal wearable approaches \cite{ni2024survey}. Combodied perception extends the problem from labeling activities to detecting changes that matter over time: a fall, a missed routine, persistent inactivity, a rehabilitation exercise, a change in gait, or a behavioral response to earlier support.

These events remain ambiguous without context. Staying at home may indicate rest, illness, remote work, poor weather, or preference, while missing data may simply mean that a device was not worn. Personal baselines, wear-state detection, environmental context, and explicit reports are therefore needed before a movement pattern becomes a claim about health or motivation. Spatial and temporal precision should also be limited to what the agreed support purpose requires; routine assistance should not become unrestricted mobility tracking.

\subsection{Social and Relational Data}

Motion traces describe what a person does, but less often explain the relationships in which those activities occur. Social evidence may come from user-described relationships, calendars, shared tasks, caregiver interactions, clinician communication, collaboration systems, communication metadata, proximity measurements, conversational turn-taking, or recurring group activities. Collection may be active, device-mediated, wearable, or ambient. The distinction between communication content and metadata is important: timing and duration can reveal a pattern without granting access to what was said, although even metadata may expose sensitive relationships.

Early mobile-sensing work showed how longitudinal phone and proximity data can reveal recurring social and organizational patterns \cite{eagle2006reality}. StudentLife combined smartphone sensing and self-report to study changes in activity, sociability, sleep, and wellbeing in a natural setting \cite{wang2014studentlife}. Sotopia approaches social interaction from a different direction, evaluating role- and goal-dependent behavior in simulated situations \cite{zhou2024sotopia}. Together, these research lines highlight different parts of the problem, but none makes relationship inference straightforward. Proximity does not necessarily imply meaningful interaction, and reduced communication does not necessarily indicate a deteriorating relationship.

The more fundamental difficulty is that relational data rarely belong to only one person. A user may authorize access to a shared calendar without authorizing the agent to infer another participant's emotional state or store their private information. Social event records therefore need to identify the focal user, other participants, the source of the relationship claim, and the authority associated with each role. Shared memories may require participant-specific visibility and correction rather than a single permission attached to the user who operates the agent.

\subsection{Environmental and Contextual Data}

Environmental context often matters not because it is itself the target of inference, but because it changes the interpretation of other evidence. Light, noise, temperature, humidity, air quality, occupancy, room state, weather, traffic, calendar information, device state, and connectivity can explain why behavior or sensor measurements have changed. These data arrive through fixed environmental sensors, phones, vehicles, smart-home infrastructure, or external services. Some are continuously sensed, while others are retrieved only when an event needs to be interpreted.

Context-aware computing has long studied the conversion of sensor readings into descriptions of situation and activity. AWARE provides an extensible framework for mobile sensing and experience sampling \cite{ferreira2015aware}, and systems such as StudentLife illustrate how phone sensing can be combined with longitudinal self-report outside the laboratory \cite{wang2014studentlife}. In a companion setting, contextual events may include arriving at work, entering a noisy environment, beginning a journey, losing connectivity, or experiencing poor air quality. Their primary value is often disambiguation: high temperature may help explain an elevated heart rate, travel may explain interrupted sleep, and a public setting may make an otherwise appropriate intervention poorly timed.

Context becomes personal when linked to identity and routine. Precise location histories can reveal home, work, health visits, religious practice, and relationships even when no explicit personal label is collected. Coarse, temporary, or event-triggered context may therefore be preferable to persistent reconstruction of the user's movements.

\subsection{Clinical, Institutional, and Structured Records}

Structured records differ from sensor streams in both temporality and authority. Diagnoses, medication lists, laboratory results, treatment plans, appointments, educational records, workplace schedules, evaluations, financial obligations, and legal documents are usually updated episodically. They may be supplied by the user, extracted from uploaded documents, synchronized periodically, received through an institutional event feed, or retrieved through an authorized query. In healthcare, HL7 FHIR and SMART on FHIR support interoperable access to clinical resources and application context \cite{mandel2016smartfhir}, while common data models such as OMOP help normalize heterogeneous databases \cite{voss2015omop}.

These records can anchor events that are difficult to infer from sensing alone: a medication change, a new diagnosis, a completed laboratory test, a missed appointment, an educational deadline, or a formal change in responsibility. MIMIC-IV illustrates the diversity of measurements, orders, diagnoses, procedures, treatments, and notes that may coexist within an electronic health record \cite{johnson2023mimiciv}. Med-PaLM investigates reasoning over medical knowledge \cite{singhal2023large}, while BEHRT and Med-BERT model temporally ordered clinical records \cite{li2020behrt,rasmy2021medbert}. These resources are important precedents, although institutional data used by a real companion will often be narrower, permission-scoped, and updated at irregular intervals.

Authority should not be confused with completeness or timeliness. A record has at least three relevant times: when the underlying event occurred, when it was entered, and when the agent received it. Corrections and version changes may alter its meaning. The agent should preserve the issuing organization, responsible professional where applicable, coding system, access scope, version, and review date. Access to a clinical, legal, financial, or employment record also does not transfer the corresponding professional authority to the agent.

Table~\ref{tab:personal-data-modalities} summarizes the main acquisition configurations. It is intentionally descriptive rather than prescriptive: listing a sensor or interface does not imply that its use is necessary or authorized.

\begin{table*}[t]
\centering
\caption{Representative acquisition configurations for personal data modalities. Acquisition mode affects coverage, burden, privacy, and interpretation; it does not imply permission for unrestricted collection.}
\label{tab:personal-data-modalities}
\footnotesize
\setlength{\tabcolsep}{3.5pt}
\renewcommand{\arraystretch}{1.05}
\begin{tabularx}{\textwidth}{
    >{\raggedright\arraybackslash}p{0.11\textwidth}
    >{\raggedright\arraybackslash}p{0.25\textwidth}
    >{\raggedright\arraybackslash}p{0.19\textwidth}
    >{\raggedright\arraybackslash}p{0.17\textwidth}
    >{\raggedright\arraybackslash}X}
\toprule
\textbf{Modality}
& \textbf{Sensors / interfaces}
& \textbf{Acquisition mode}
& \textbf{Sampling}
& \textbf{Event focus} \\
\midrule

Language / text
& Keyboard, touchscreen, speech transcription, OCR, application interfaces, uploaded documents
& User-reported, prompted, or device-mediated
& User-initiated, episodic, periodic, or context-triggered
& Goals, commitments, preferences, corrections, self-reports, and consent changes. \\

Speech / audio
& Device microphones, earbuds, hearing devices, smart speakers, contact microphones, microphone arrays
& Near-field wearable, device-mediated, or ambient contactless
& Conversation-bound, voice-triggered, event-triggered, or continuous
& Speech episodes, turn-taking, acoustic change, and safety-relevant sounds. \\

Vision
& RGB, stereo, depth, infrared, thermal, eye-tracking, and event cameras
& Device-facing, wearable egocentric, in-vehicle, or fixed ambient
& Interaction-bound, periodic, event-triggered, or continuous
& Activity, gaze, posture, object interaction, social episodes, and scene changes. \\

Physiology / biochemistry
& ECG, PPG, SpO$_2$, EDA, respiration, temperature, blood pressure, glucose, and biochemical sensors
& Wearable contact, skin patch, minimally invasive, clinical, or contactless
& Continuous waveform, periodic estimate, episodic, or threshold-triggered
& Baseline deviation, recovery, sustained trend, and acute anomaly. \\

Motion / behavior
& Inertial sensors, GPS, BLE, Wi-Fi, UWB, PIR, pressure sensors, smart objects, and radar
& Phone-carried, body-worn, object-embedded, or ambient
& Continuous, periodic, state-transition, or event-triggered
& Activity episodes, mobility changes, falls, routines, adherence, and response. \\

Social / relational
& Communication metadata, calendars, collaboration logs, proximity sensors, audio, and vision
& User-reported, device-mediated, wearable proximity, or ambient multi-party
& Episodic, interaction-triggered, periodic aggregation, or continuous
& Relationship and role changes, coordination, withdrawal, conflict, and shared obligations. \\

Environment / context
& Light, noise, temperature, air quality, occupancy, smart-home state, weather, traffic, and maps
& Fixed ambient, mobile, infrastructure-mediated, or external service
& Continuous, periodic, state-transition, context-triggered, or on demand
& Situation changes, exposure, contextual explanation, and intervention timing. \\

Structured records
& FHIR/HL7 interfaces, databases, event feeds, batch exports, user uploads, and OCR
& Institutional push, authorized query, periodic synchronization, or user-mediated import
& Episodic, transactional, periodic, or on demand
& Diagnoses, plans, obligations, appointments, corrections, and formal constraints. \\

\bottomrule
\end{tabularx}
\end{table*}

\subsection{Data Quality, Provenance, and Uncertainty}

The modalities above differ not only in content but also in reliability. A heart-rate estimate from a loose watch, speech recorded in a noisy room, video captured under poor lighting, a self-report written during acute distress, and a clinical record entered several weeks after an encounter cannot be treated as equivalent evidence. Uncertainty may enter during acquisition, subject attribution, event segmentation, interpretation, or later use. Collapsing these sources into a single confidence score would make it difficult to identify why an event is unreliable.

An event-evidence record should retain enough acquisition metadata to support later review: observation and acquisition times, source, device or sensor type, placement where relevant, sampling mode, calibration or software version, and available signal-quality indicators. Processing provenance should identify transformations such as denoising, compression, transcription, feature extraction, event segmentation, and the model version that produced an inference. The interpretation itself should distinguish observed fields from inferred fields, preserve alternative explanations and contradictory evidence, and indicate the personal baseline against which a deviation was judged. Finally, sensitivity, consent basis, use scope, third-party involvement, retention, visibility, correction history, and deletion conditions determine how the record may be used. These requirements extend documentation practices such as Data Cards \cite{pushkarna2022datacards} from datasets to longitudinal personal evidence.

Quality control is not a single preprocessing step. At acquisition time, the system may reject or down-weight data affected by poor contact, non-wear, occlusion, noise, calibration failure, or uncertain identity. Temporal alignment must then account for clock drift, asynchronous sampling, uncertain event boundaries, and delayed records. During interpretation, the system should preserve ambiguity rather than force every fragment into a state label. Memory writing adds another gate: even a reliable event may be too sensitive, temporary, irrelevant, or weakly consented to retain.

Missingness requires similar care. A gap in wearable data may result from depleted battery, device removal, connectivity failure, refusal, or an actual change in routine. Missing data can become evidence only when the acquisition process makes that interpretation plausible. Likewise, a model's statistical confidence does not establish sensor reliability, causal validity, or permission to intervene.

The purpose of provenance is not to preserve raw personal data indefinitely. Raw audio, video, or high-frequency physiological signals may be deleted after local processing while a constrained event description and its quality indicators are retained. Users should still be able to inspect what was observed, what was inferred, which sources supported the interpretation, what contradictory evidence existed, and how the event affected memory or action. The retained provenance must itself remain subject to consent and deletion.

\subsection{Event-Based Multimodal Fusion and Evidence Reconstruction}

Multimodal fusion \cite{liang2022foundations} is often described as combining representations or predictions from several sources. For Combodied Agents, the harder problem is reconstructing an evidence chain from observations that are sparse, delayed, and only partially overlapping. Fusion begins with candidate events produced within individual modalities. It then asks whether several fragments plausibly refer to the same episode, whether they corroborate or contradict one another, and whether the resulting interpretation is relevant to the person's goals or safety.

This process includes fragment filtering, temporal alignment, cross-modal comparison, and reconstruction of event context. Fragment filtering reduces long streams to segments that are potentially informative. Temporal alignment connects evidence before, during, and after an event while respecting uncertain boundaries and different sampling rates. Cross-modal comparison determines whether language, speech, vision, physiology, motion, social, environmental, and institutional evidence offer compatible explanations. Reconstruction links the event to personal baselines, relevant memories, previous interventions, and governance constraints.

Consider an elevated heart-rate measurement. It should first remain a physiological observation rather than be labeled as stress or illness. Motion data may indicate that it occurred during exercise; weather data may show high temperature; a later self-report may mention fatigue; and an intervention-response record may show that the user postponed an earlier rest suggestion. Together, these fragments may support an event describing elevated exertion and delayed recovery. They still do not establish a diagnosis or automatically authorize action. If important alternatives remain unresolved, clarification may be a better perceptual outcome than a stronger inference.

Cross-modal agreement is also not automatically correct. Several sensors may share a common failure, and one high-quality user correction may outweigh multiple indirect signals. Fusion therefore needs to be person-calibrated and provenance-aware rather than based only on the number of agreeing modalities. It should also be purpose-limited: once the agreed support objective has sufficient evidence, additional sensing may increase privacy risk without improving the decision. Sensitive audio, vision, physiology, location, and relationship data should be processed locally where feasible \cite{chung2026localprivacy}, with raw retention determined by modality-specific risk and explicit user control.

Most importantly, the architecture must distinguish an observation from an event, an event from an inferred state, an inferred state from a predicted trajectory, and a predicted trajectory from permission to intervene. Text, physiological signals, speech, and vision are more precisely sources or modalities of event evidence rather than events by themselves. For example, an elevated heart rate is an observation; elevated heart rate with slow recovery after exercise is a reconstructed event; possible fatigue is an inferred state; further delayed recovery over the next hour if exercise continues is a predicted trajectory; and recommending that the person stop exercising is a decision made by the Intervention Policy. The distinction can be summarized as:

\begin{center}
\small
\textit{observation} $\rightarrow$ \textit{event} $\rightarrow$ \textit{inferred state} $\rightarrow$ \textit{predicted trajectory} $\rightarrow$ \textit{authorized intervention}.
\end{center}

The downstream interface is therefore the governed event-evidence record defined in this section, not a general-purpose stream of personal data. Section~\ref{sec:pwm} treats such records as uncertain evidence for latent personal state and personal dynamics. Whether the agent may act on a predicted state or trajectory remains a separate question for intervention policy, consent, proportionality, and safety.

\section{Personal World Model}
\label{sec:pwm}

\subsection[Definition and Positioning]{Definition and Positioning}

A world model is an internal predictive representation of how a relevant environment or state evolves, particularly under actions. By supporting imagined future trajectories, it enables an agent to anticipate consequences, plan, and control without testing every action directly in the real environment \cite{Ha2018WorldModels,Schrittwieser2019MuZero,Hafner2023DreamerV3}.

A personal model is an updateable representation of a particular individual's current condition, characteristics, goals, routines, capabilities, relationships, constraints, and history. We define a \textit{Personal World Model} (PWM) more specifically as \textbf{a purpose-bounded, individual-specific event-dynamics model}. Given a governed history of multimodal event-evidence records for a particular person, the current spatiotemporal context, and a candidate scenario specifying user decisions, agent interventions or support, and relevant environmental changes, a PWM assimilates the event history into an uncertainty-bearing representation of the person's current state and predicts a calibrated distribution over future human states, observable personal events, and scenario-relevant outcomes.

The defining function of a PWM is therefore not personalization or plausible behavior generation alone, but intervention-conditioned modeling of how this particular person's state--event trajectory may unfold under alternative scenarios. Subsequent events and observed intervention responses update its representation of personal dynamics. A PWM supports scenario comparison, but it does not itself select or authorize an intervention; action selection remains the responsibility of the Intervention Policy under consent, safety, reversibility, and agency-preservation constraints.

A PWM may use profiles, longitudinal memory, personalized foundation models, generative simulation, causal models, or mechanistic models as implementation components. These components become part of a PWM only when they jointly support person-specific, context-sensitive, and action-conditioned prediction of future state--event trajectories with explicit uncertainty. A PWM is consequently a family of purpose- and horizon-specific models, not a monolithic simulation of a whole person. Table~\ref{tab:pwm-related-constructs} positions this functional contract relative to profiles, memory, personalized agents \cite{Xu2026PersonalizedAgents}, and generative agents \cite{park2023generative}.

\begin{table}[H]
\centering
\caption{Positioning PWMs relative to adjacent personal representations and agent-modeling constructs. The constructs may overlap; the distinction concerns their functional contracts rather than an exclusive capability boundary.}
\label{tab:pwm-related-constructs}
\small
\renewcommand{\arraystretch}{1.2}
\setlength{\tabcolsep}{3pt}
\begin{tabularx}{\textwidth}{>{\raggedright\arraybackslash}p{0.13\textwidth}>{\raggedright\arraybackslash}p{0.13\textwidth}>{\raggedright\arraybackslash}p{0.24\textwidth}>{\raggedright\arraybackslash}p{0.25\textwidth}>{\raggedright\arraybackslash}X}
\toprule
\textbf{Construct} & \textbf{Conceptual level} & \textbf{Central question} & \textbf{Scenario-conditioned output} & \textbf{Defining validation} \\
\midrule
User profile & Representation & What attributes or preferences characterize the user? & Usually none; supports conditional personalization & Attribute and preference accuracy \\
Longitudinal memory & Memory component & What has happened to the person? & Retrieval of events, goals, corrections, and prior responses rather than dynamics prediction & Recall, provenance, correction, retention, and deletion \\
Personalized agent & Agent system & How should outputs, plans, or actions adapt to the user? & May predict future user data, but this is not its defining contract & Personalized task performance and user benefit \\
Generative agent or model & Agent or modeling paradigm & What behavior or scenario can be plausibly generated? & May simulate future behavior, but need not model calibrated intervention responses for a particular real person & Plausibility, consistency, or simulation fidelity \\
Personal World Model & Individual dynamics-model component & How will this person's states and events evolve under alternative decisions, interventions, and contexts? & Calibrated distributions over future state--event--outcome trajectories & Personal trajectory validity, intervention-response accuracy, calibration, and scenario discrimination \\
\bottomrule
\end{tabularx}
\end{table}

These constructs are not mutually exclusive. A personalized agent may contain a PWM, and a generative model may implement part of a PWM. The distinction lies in the model's functional contract: a PWM is explicitly responsible for learning and validating the action- and context-conditioned dynamics of a particular person's state--event trajectory. Retrieving personal context, adapting an answer, or predicting the next user message does not by itself satisfy this contract.

\subsection{Predictive Dynamics and Intervention}

The state posterior and retrieved evidence are defined in Eq.~\ref{eq:state-posterior}. The governed history $D_{\leq t}$ contains multimodal event-evidence records rather than an undifferentiated raw-data stream. A modeling interface may decompose $Z_t$ into physiological, cognitive, emotional, behavioral, social, and relational components, while keeping the user's stated and inferred goals in a separate, correctable variable $G_t$; the components need not be independent or completely observed.

Let a candidate scenario
$\mathbf{s}_{t:t+\Delta}=(\mathbf{a}^{\mathrm{user}},\mathbf{a}^{\mathrm{agent}},\boldsymbol{\Xi})_{t:t+\Delta}$
specify possible user decisions, agent interventions or support, and relevant environmental or social changes. A PWM estimates a distribution over future latent states, observable personal events, and downstream outcomes:

\begin{equation}
    p_\theta\!\left(
    Z_{t+1:t+\Delta},E_{t+1:t+\Delta},Y_{t+1:t+\Delta}
    \mid D_{\leq t},Z_t,C_t,G_t,\mathbf{s}_{t:t+\Delta}
    \right),
    \label{eq:pwm-dynamics}
\end{equation}

where $E$ denotes future observable personal events and $Y$ denotes scenario-relevant outcomes such as adherence, goal progress, wellbeing, capability, safety, relationship quality, and agency preservation. Because future user responses and environmental events remain uncertain even when a candidate scenario is specified, their unresolved components are sampled, varied, or marginalized during a rollout rather than treated as known facts. Useful scenario sets include non-intervention, clarification, alternative timings or intensities of support, user acceptance or refusal, and relevant contextual changes. The defining comparison is therefore not only what happens next, but how the distribution over this person's future state--event trajectory changes across explicit alternatives.

The PWM informs but does not by itself authorize intervention. Mirroring the closed-loop framework, the policy identifies nondominated actions only within the admissible set:

\begin{equation}
    a_t^{\mathrm{agent},*}
    \in
    \underset{a \in \mathcal{A}_t^{\mathrm{adm}}}
    {\operatorname{ParetoArgmax}}
    \;\mathbb{E}_{p_\theta^{a}}
    \left[\mathbf{U}\!\left(Y_{t+1:t+\Delta},G_t\right)\right],
    \label{eq:pwm-intervention-policy}
\end{equation}

where $\mathcal{A}_t^{\mathrm{adm}}$ enforces consent, scope, safety, uncertainty, reversibility, and escalation requirements and includes non-intervention, clarification, and referral. The objective vector $\mathbf{U}$ preserves explicit trade-offs among benefit, capability, autonomy, relationships, and other scenario outcomes. A user-approved selection rule is still required among nondominated actions; safety and consent cannot be exchanged for higher engagement or average predicted utility.

\begin{figure}[t]
\centering
\includegraphics[width=\linewidth]{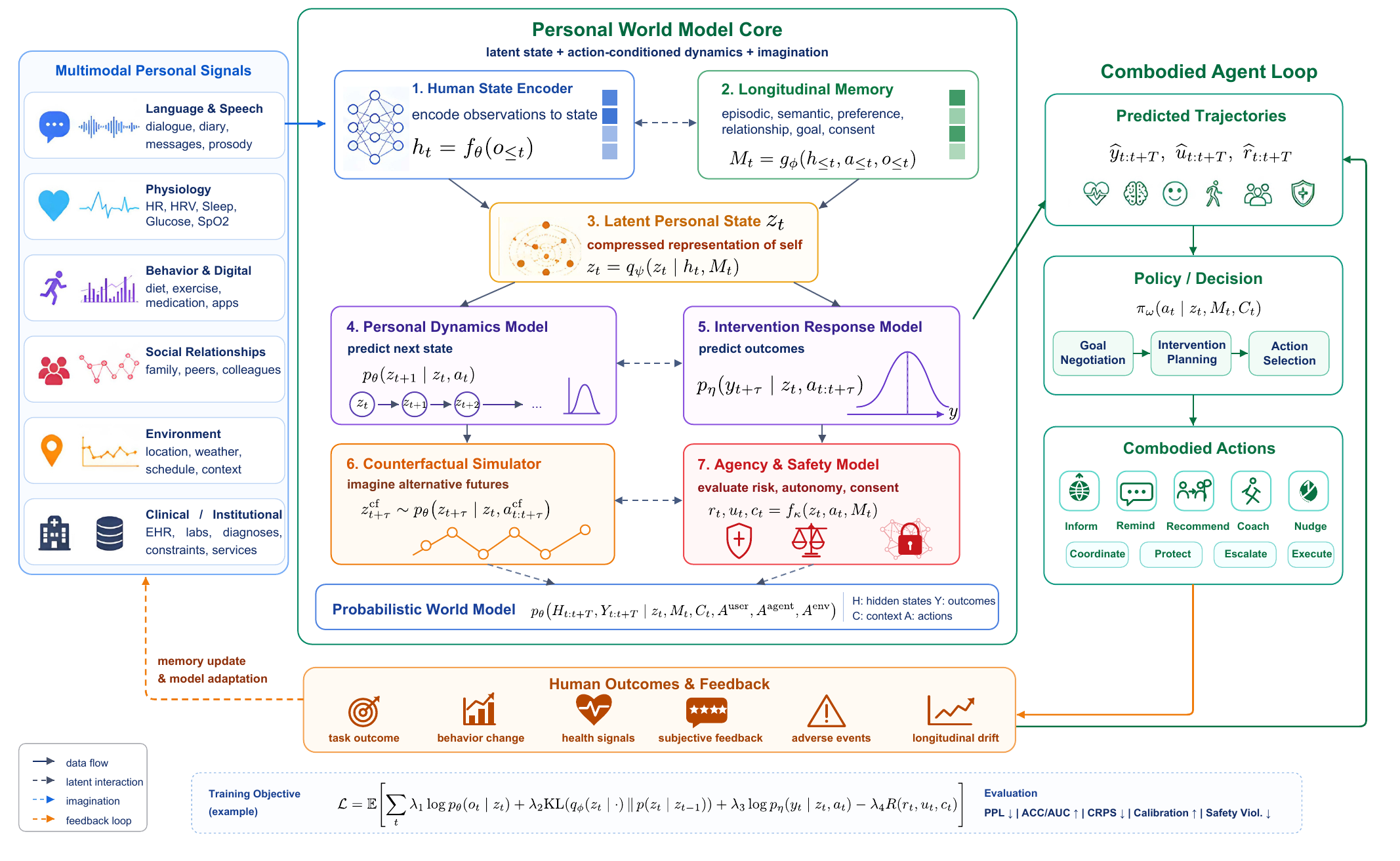}
\caption{Reference technical scheme of a PWM. Event evidence and longitudinal memory support a posterior over latent personal state; domain- and horizon-specific dynamics and response models generate state--event--outcome trajectories under alternative decisions, interventions, and uncertain user and environmental responses; and an admissibility boundary constrains decision, action, feedback, and model update.}
\label{fig:pwm-tech-scheme}
\end{figure}

Figure~\ref{fig:pwm-tech-scheme} situates the predictive model and admissibility boundary inside the larger feedback loop. Its component losses and metrics are illustrative because different domains and horizons require different estimators and validation standards.

% \clearpage
% \clearpage

Evidence-grounded state inference must preserve contradictory evidence and uncertainty; reasoning over wearable data, for example, demonstrates an encoder or reasoning interface but not by itself a learned personal dynamics model \cite{Merrill2026PHIA}. Personalized and generative agents may also predict future user data. What defines the PWM component is the explicit transition-modeling contract from a particular person's governed event history and current spatiotemporal context to calibrated state--event--outcome trajectories under alternative decisions and interventions, updated from subsequent events and intervention-response memory \cite{Lin2026LongitudinalHealth}.

A predictive PWM compares scenario distributions under explicit assumptions. A stronger causal PWM may additionally support interventional queries of the form

\begin{equation}
    p_\theta\!\left(Y_{t+1:t+\Delta}
    \mid \operatorname{do}(a_t^{\mathrm{agent}}=a),Z_t,R_t,C_t\right).
    \label{eq:pwm-interventional-query}
\end{equation}

This is an interventional distribution, not automatically an identified individual counterfactual. A statement about what would have happened to the same person under $a'$ requires potential outcomes such as $Y_t(a)$ and $Y_t(a')$, or a structural causal model with an explicit abduction--action--prediction procedure. Observed user behavior is confounded by motivation, hidden context, health status, prior interactions, and selective engagement. Writing $\operatorname{do}(\cdot)$ does not remove those confounders. Identification requires a defensible causal graph and assumptions such as consistency, positivity, and appropriate control of time-varying confounding, supported where possible by randomized or micro-randomized interventions, N-of-1 studies, or carefully validated observational estimators. High-risk systems must not conduct unconstrained exploration merely to improve the model.

Agency-and-safety governance remains outside the predictive model: a high predicted benefit is neither a factual guarantee nor permission to act. Model scope, consent, roles, reversibility, and escalation determine the admissible set and which data may be retained for adaptation.

\subsection{Learning Paradigms, Fidelity, and Limits}

Methods for PWMs can be organized into several paradigms. \textit{Predictive latent dynamics} follows the model-based reinforcement-learning tradition: encode human observations into a latent personal state and learn transitions under user actions and agent interventions. \textit{Generative scenario simulation} follows the trajectory of generative world models, producing structured future scenarios rather than only scalar predictions \cite{Ding2024WorldModelSurvey}. \textit{Causal intervention modeling} estimates treatment or intervention effects and supports counterfactual reasoning. \textit{Mechanistic and hybrid personal dynamics modeling} combines validated domain knowledge with learned dynamics when physiological, behavioral, or environmental structure is available. \textit{Mental-state modeling} represents beliefs, desires, intentions, emotions, trust, and relationship dynamics. \textit{Memory-augmented modeling} uses structured profiles, event timelines, temporal knowledge graphs, and retrieval-augmented generation. \textit{Hybrid foundation-personal modeling} combines population-level priors with individual adaptation layers, personal memory, causal modules, and safety constraints.

Sparse personal data make training a model from scratch inappropriate for most users. A practical PWM should begin with population-level or domain-level priors, adapt a limited set of personal components, maintain a posterior over uncertain parameters or states, detect personal and environmental drift, and support correction or reset. Different time scales should use hierarchical or separate models: a short-horizon fatigue predictor, a medium-horizon habit model, and a long-horizon capability assessment need not share the same state, loss, or validation threshold. Long-horizon outputs should be treated as scenarios with widening uncertainty, not precise forecasts of a life trajectory.

PWMs can therefore be categorized by human-state target, time horizon, intervention type, modeling paradigm, evidence quality, and application domain. Their targets may be physiological, cognitive, emotional, behavioral, social, relational, goal-directed, or deliberately restricted combinations. Their time horizons may range from a moment or day to weeks or months; year- or lifespan-level projections should be framed as exploratory scenarios unless supported by unusually strong longitudinal evidence. Their intervention types may include informing, reminding, recommending, coaching, nudging, reflecting, coordinating, protecting, escalating, and executing. Their application domains include health, education, productivity, emotional support, eldercare, and guardian systems.

Different Combodied Agents require different model fidelity and abstention thresholds. A medication-adherence companion needs calibrated risk estimates, clinical boundaries, and escalation rules; a productivity companion may use a lower-fidelity preference and schedule model; and a romantic or emotional companion requires robust relationship-safety modeling because prediction errors can intensify dependency or manipulation. AI companion studies show that usage patterns and user characteristics shape psychosocial outcomes, including loneliness and problematic use \cite{Liu2024ChatbotCompanionship,Ciriello2026Loneliness}. Reports on teen AI companion use and safety further motivate age-sensitive safeguards and privacy controls \cite{CommonSense2025Companions}, while supervisory systems and persona-grounded evaluations begin to test over-attachment, isolation reinforcement, and boundary violations \cite{BenZion2025SHIELD,Juneja2026PersonaSafety}. These findings constrain the fidelity claims and authority of a PWM; they are not merely downstream evaluation concerns.

PWM is thus an organizing abstraction rather than a single newly claimed estimator. Its research contribution lies in connecting person-specific predictive dynamics to a correctable, agency-constrained intervention loop. Its empirical validity must be established separately for each target, horizon, population, and authority level. Because the most sensitive state, memory, and intervention history should remain inspectable and controllable by the user, the next section examines how this model family can be partitioned across trusted edge devices and selectively invoked cloud services.

\section{From Cloud LLMs to Edge Personal Models}
\label{sec:edge-personal-models}

\subsection[Edge-Native Personal Models]{Edge-Native Personal Models}

Because Combodied Agents maintain sensitive longitudinal models, their core personal intelligence should be local-first on trusted devices \cite{shi2016edge,zhou2019edgeintelligence,deng2019edgeintelligence,tian2026edgeagents}. We define an \textit{edge-native personal model} as a user-controlled stack that maintains longitudinal memory, the PWM, preferences, intervention policy, and safety boundaries primarily on user-side devices. Local ownership supports privacy, continuity, low latency, resilience, and inspection or reset. It does not imply a cloud-free system: external models may supply knowledge, specialized expertise, or computation, while the edge controls disclosure and interprets returned results.

\subsection[Cloud-to-Edge Evolution]{Cloud-to-Edge Evolution}

Combodied-Agent deployment can be organized into three stages according to where personal memory, human-state interpretation, reasoning, and intervention authority reside. Stage I is cloud-centric, Stage II uses the edge to mediate personal context and cloud capability, and Stage III makes a user-controlled edge model the primary locus of personal intelligence.

These stages describe architectural centers of gravity rather than a universal or strictly chronological product sequence. A single Combodied Agent may operate at different stages for different tasks: routine reminders and private memory retrieval may be edge-native, specialized planning may use a hybrid pipeline, and broad knowledge queries may remain cloud-centric. Progress between stages should therefore be determined not only by improvements in local computation, but also by requirements for privacy, latency, resilience, model correctability, and control over intervention authority. The central migration question is where the authoritative representation of the person and the final authority to act should reside for a given task and risk level.

\subsubsection{Stage I: Cloud-Centric Combodied Agents}

% In the first stage, Combodied Agents are built primarily on existing cloud-hosted large language models. User inputs, contextual information, and relevant memory are sent to the cloud model, which performs perception, reasoning, planning, and response generation. Personalization is usually implemented through prompting, retrieval-augmented memory, cloud-side user profiles, or tool calling.

% This stage benefits from the strongest available foundation models and rapid product iteration. It is suitable for early Combodied Agent systems that require broad language ability, general reasoning, and flexible interaction. However, it has substantial limitations. Sensitive personal data often leave the user's device; longitudinal memory is controlled by the service provider rather than the user; personalization is dependent on cloud-side policies; and continuous model evolution around the individual is difficult to make transparent, reversible, or portable.

% In this stage, the Combodied Agent is best understood as a \textit{cloud assistant with personal context}. It may remember, recommend, and support, but its personal intelligence is not yet locally grounded or user-owned.

In Stage I, a cloud-hosted foundation model performs most reasoning, memory retrieval, tool selection, and response generation, while the user device functions mainly as an interface, sensor endpoint, and execution surface. Personalization is typically implemented through prompting, retrieved dialogue histories, profiles, preferences, or episodic records. The user's state is therefore reconstructed from the context supplied to each request rather than maintained as an explicit, continuously updated PWM under the user's control.

This stage remains valuable for rapid prototyping, cold-start interaction, and low-frequency or low-risk applications that require broad knowledge, flexible language understanding, or access to powerful external tools. It allows new Combodied-Agent scenarios to be evaluated before sufficiently capable local models and personal data have been established. Cloud-centric deployment is therefore not merely an incomplete version of edge-native intelligence; it is a practical architecture when the personal context is limited, the task is reversible, and the claimed intervention authority is narrow.

Its technical and governance boundary arises when retrieved context is treated as if it were a persistent personal model. A cloud-side profile or memory store does not by itself provide a user-owned PWM, and service-controlled memory may be difficult to inspect, migrate, or correct consistently. Continuous sensing and high-impact interventions further increase the consequences of sending raw personal evidence to a remote service. Cloud-generated outputs should therefore not directly authorize irreversible actions solely on the basis of inferred personal state. High-impact interventions require explicit confirmation or another trusted decision boundary, and user correction or deletion requests must propagate to every representation that may affect subsequent behavior.

Migration toward Stage II becomes warranted when sensitive multimodal data would otherwise be continuously uploaded; when the agent requires an authoritative and correctable cross-session memory; when latency, offline operation, or service resilience becomes important; or when users require practical control over the inspection, deletion, export, and portability of the model state that represents them. Empirically, Stage I systems should be evaluated by how accurately retrieved context approximates longitudinal state, how reliably corrections affect future interactions, how much sensitive information is disclosed per task, and how memory continuity degrades under network failure, service change, or incomplete context retrieval.

% \subsubsection{Stage II: Hybrid Edge-Cloud Combodied Agents}

% In the second stage, Combodied Agents adopt a hybrid edge-cloud architecture. The edge device performs privacy-sensitive perception, local memory filtering, short-term context tracking, safety checks, and lightweight personalization, while the cloud is invoked for complex reasoning, broad knowledge retrieval, large-scale generation, or specialized tasks that exceed local computational capacity.

% In this architecture, the edge-resident system acts as a personal privacy and interpretation layer. It determines what information remains local, what information can be summarized, and what sanitized query may be sent to the cloud. The cloud model no longer receives raw personal memory by default. Instead, it receives purpose-limited abstractions, user-approved summaries, or de-identified task representations.

% This stage is transitional but highly practical. It allows Combodied Agents to benefit from cloud-scale intelligence while gradually moving personal state modeling, longitudinal memory, and intervention adaptation closer to the user. It also introduces an important architectural principle:

% \begin{quote}
% The cloud may assist the Combodied Agent, but the edge should mediate the user's personal world.
% \end{quote}

\subsubsection{Stage II: Hybrid Edge-Cloud Combodied Agents}

In Stage II, the edge becomes a privacy, interpretation, and authority mediator rather than a passive interface. It performs privacy-sensitive perception, local context tracking, memory filtering, safety checks, and task routing, while cloud services remain available for external knowledge, complex reasoning, large-scale generation, and specialized tools \cite{chung2026localprivacy}. Raw signals and private memories can remain local, with the cloud receiving purpose-limited summaries or abstracted task representations.

A system should not be classified as Stage II merely because it runs a small model on the device. At minimum, the edge should maintain an authoritative copy of sensitive personal state or memory, enforce disclosure and action permissions, record the information sent to external services, and review cloud results before they update protected memory or trigger consequential action. Cloud assistance may contribute reasoning, but it should not silently redefine user goals, overwrite local corrections, or bypass the local intervention policy.

The central technical boundary in Stage II is the division of state and authority across heterogeneous components. Task routing can misclassify sensitivity or computational need; supposedly sanitized representations may still reveal identity, health, relationships, or routines; and edge and cloud components may hold inconsistent versions of personal context. Model updates in the cloud can also change how the same local summary is interpreted. A hybrid system must therefore specify which representation is authoritative, how provenance is preserved across calls, how conflicts are resolved, and which functions remain available or safely degrade when connectivity is lost.

Migration toward Stage III becomes justified when the edge can perform most recurrent personal tasks at an acceptable and calibrated quality; maintain and update memory, the PWM, and intervention constraints without reconstructing the user in the cloud; support inspection and rollback of local adaptation; and preserve essential safety and control during disconnection. At this point, cloud use changes from the default reasoning path to an explicit exception for tasks that exceed local knowledge or computational capacity.

Stage II can be evaluated through controlled comparisons with cloud-only and local-only baselines. Relevant questions include whether routing policies achieve a measurable privacy--utility--latency trade-off; how much personal information can be removed without degrading task outcomes; how often local and cloud states conflict; whether cloud outputs are appropriately rejected, revised, or escalated by the edge; and whether the system fails safely under network interruption, stale cloud models, or incomplete audit trails. These tests make hybrid deployment a verifiable architectural claim rather than a generic description of distributed computation.

\subsubsection{Stage III: Edge-Native Personal Combodied Agents}

% In the third stage, personal tasks are primarily handled on the edge. The user's PWM, longitudinal memory, intervention policy, preference model, and personal safety boundaries are stored and updated on trusted user-side devices. The agent's evolution also occurs locally: it adapts its memory, state perception, personal dynamics, communication style, intervention responses, and boundary conditions based on the user's longitudinal feedback.

% In this stage, the cloud is used selectively rather than continuously. For ordinary personal tasks, such as interpreting local routines, retrieving personal memory, selecting reminders, adapting tone, or updating preferences, local inference is sufficient. For complex thinking tasks that require broad external knowledge, scientific literature, specialized medical or legal information, or large-scale reasoning, the edge-resident agent may invoke a cloud LLM or external retrieval system.

% The Stage III architecture can be described as \textit{local by default and cloud by exception}. The edge personal model becomes the user's primary combodied intelligence, while cloud foundation models become external cognitive tools used only when necessary.

In Stage III, the authoritative copies of longitudinal memory, the PWM, preferences, intervention policy, and personal safety boundaries are stored and updated primarily on trusted user-side devices. Requests are interpreted locally in relation to personal state and previous outcomes, and cloud services are invoked selectively for external knowledge, specialized expertise, or computation beyond local capacity. Cloud results return to the edge for contextualization and authorization before they affect protected memory or action.

The defining property of Stage III is therefore not that every computation occurs locally, but that personal interpretation and intervention authority remain under user-side control. A Stage III system should allow the user to inspect, correct, delete, pause, export, reset, and migrate the model state that represents them. Local adaptation should also be versioned and reversible, with mechanisms for detecting drift, preserving explicit user corrections, and preventing temporary states or erroneous memories from becoming persistent policy.

Edge-native deployment nevertheless has important technical boundaries. Local models face constraints in computation, storage, energy, model freshness, and access to specialized knowledge. Local adaptation may overfit short-term behavior, reinforce incorrect inferences, or produce catastrophic forgetting. Device compromise, loss, shared-device use, and inconsistent multi-device synchronization can threaten the authoritative personal state. Moreover, local storage does not automatically guarantee privacy, meaningful consent, or agency preservation. Medical, legal, crisis-related, and other high-stakes tasks may still require qualified humans or governed external services even when the personal model is edge-native.

Stage III should consequently be treated as a task-dependent target architecture rather than a universal endpoint. Its empirical evaluation should test whether local personalization improves longitudinal outcomes relative to static edge and cloud-personalized baselines; whether user corrections propagate through memory, prediction, and policy; whether unsafe local updates can be identified and rolled back; and whether the personal model remains semantically consistent after device or provider migration. Additional tests should measure performance during disconnection, energy and latency under continuous operation, recovery after device loss or synchronization conflict, and the calibration of thresholds that trigger cloud assistance or human escalation.

\begin{figure}[t]
\centering
\includegraphics[width=\linewidth]{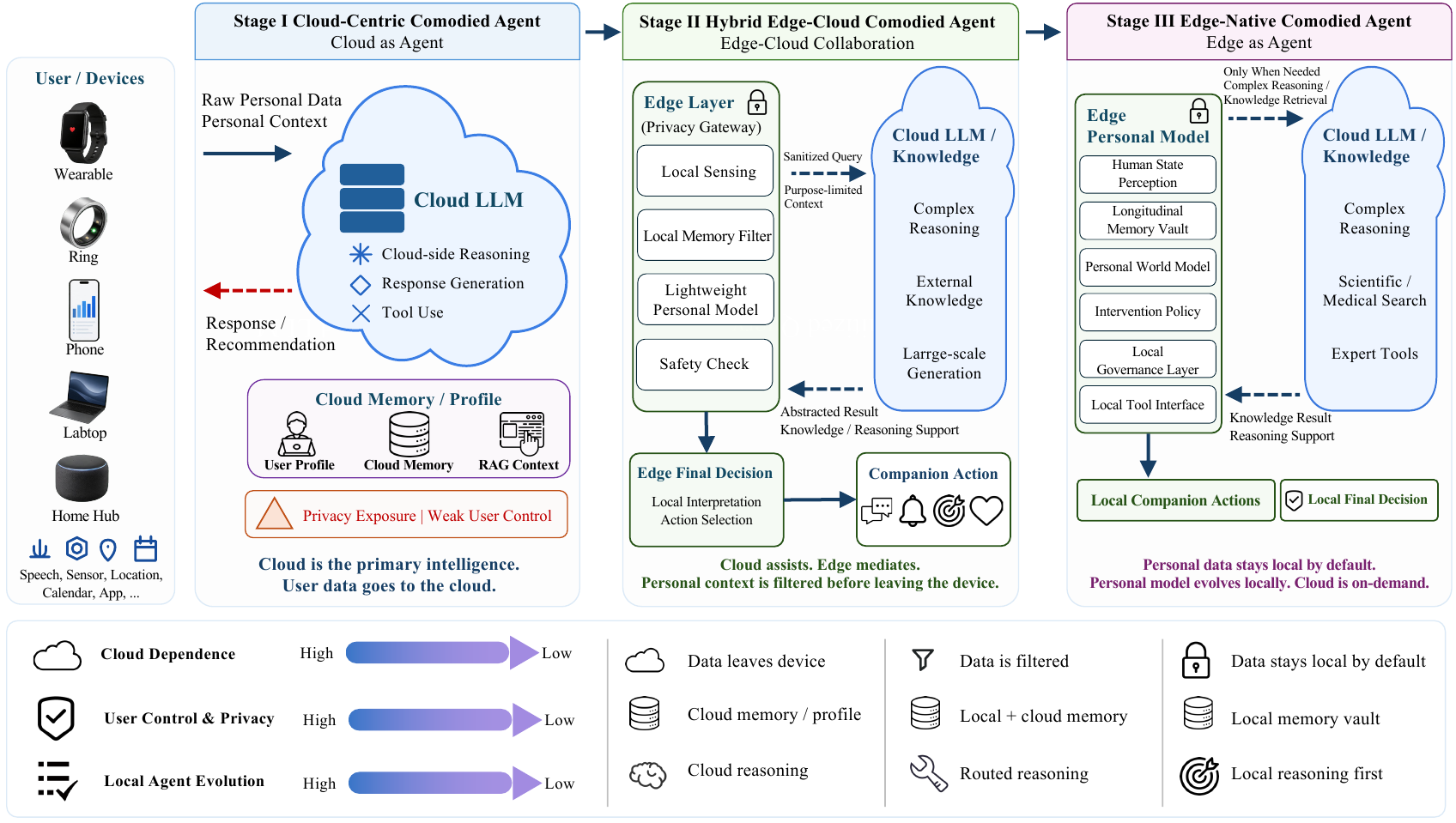}
\caption{Three-stage development of Combodied Agent deployment. The trajectory moves from cloud-centric assistants with personal context, to hybrid edge-cloud systems that mediate sensitive context locally, and finally to edge-native personal models where memory, personalization, and intervention policy primarily reside on trusted user-side devices.}
\label{fig:combodied-agent-three-stage}
\end{figure}

Figure~\ref{fig:combodied-agent-three-stage} summarizes this progression. Across the three stages, the primary transition is from cloud access to personal context, through edge-mediated disclosure and authority, toward user-controlled personal intelligence. The stages should ultimately be compared by demonstrated control, correctability, resilience, and human benefit rather than by the nominal location of model inference alone.

% \clearpage
% \clearpage

\subsection[Architecture and Routing]{Architecture and Routing}

\begin{figure}[t]
\centering
\begin{tikzpicture}[
    font=\small,
    >=Latex,
    node distance=0.75cm and 1.0cm,
    box/.style={draw, rounded corners=3pt, align=center, minimum width=3.0cm, minimum height=0.82cm},
    smallbox/.style={draw, rounded corners=3pt, align=center, minimum width=2.7cm, minimum height=0.75cm},
    group/.style={draw, rounded corners=5pt, inner sep=0.25cm, dashed},
    arrow/.style={->, thick}
]

\node[box, fill=blue!8] (signals) {Human and Contextual\\Signals};
\node[box, fill=blue!8, below=of signals] (perception) {Local Human State\\Perception};
\node[box, fill=purple!8, below=of perception] (memory) {Local Longitudinal\\Memory Vault};
\node[box, fill=purple!8, below=of memory] (pwm) {Local Personal\\World Model};
\node[box, fill=orange!10, below=of pwm] (policy) {Local Intervention\\Policy};
\node[box, fill=orange!10, below=of policy] (action) {Combodied Actions\\and Local Tools};

\node[smallbox, fill=gray!10, right=1.5cm of memory] (gateway) {Privacy Gateway\\and Context Filter};
\node[smallbox, fill=gray!10, below=of gateway] (router) {Task Router\\Sensitivity Filter};

\node[box, fill=green!8, right=1.7cm of router] (cloudllm) {Cloud LLM\\Complex Reasoning};
\node[box, fill=green!8, above=of cloudllm] (cloudret) {External Knowledge\\Retrieval};
\node[box, fill=green!8, below=of cloudllm] (cloudtools) {Specialized Cloud\\Tools};

\node[group, fit=(perception)(memory)(pwm)(policy)(action), label={[font=\small, xshift=-2.3cm]above:Edge Personal Intelligence Stack}] (edgegroup) {};
\node[group, fit=(cloudret)(cloudllm)(cloudtools), label={[font=\small]above:Cloud Assistance Layer}] (cloudgroup) {};

\draw[arrow] (signals) -- (perception);
\draw[arrow] (perception) -- (memory);
\draw[arrow] (memory) -- (pwm);
\draw[arrow] (pwm) -- (policy);
\draw[arrow] (policy) -- (action);

\draw[arrow] (memory) -- (gateway);
\draw[arrow] (pwm) -- (router);
\draw[arrow] (policy) -- (router);
\draw[arrow] (gateway) -- (router);

\draw[arrow] (router) -- node[above, font=\scriptsize] {sanitized query} (cloudllm);
\draw[arrow] (router) -- (cloudret);
\draw[arrow] (router) -- (cloudtools);

\draw[arrow] (cloudllm.west) -- node[below, yshift=-2pt, rotate=20, font=\scriptsize] {abstracted result} (policy.east);
\draw[arrow] (cloudret.west) -- (pwm.east);
\draw[arrow] (cloudtools.west) -- (action.east);

\draw[arrow] (action.west) .. controls +(-1.0,-0.2) and +(-1.0,-0.2) .. node[left, font=\scriptsize] {feedback} (memory.west);

\end{tikzpicture}
\caption{Reference architecture of an edge-native Combodied Agent. Personal state perception, longitudinal memory, the PWM, and the Intervention Policy primarily reside on the user's edge devices. Cloud models are invoked only through a privacy gateway and task router for complex reasoning, external knowledge retrieval, or specialized tools.}
\label{fig:edge-native-companion-architecture}
\end{figure}
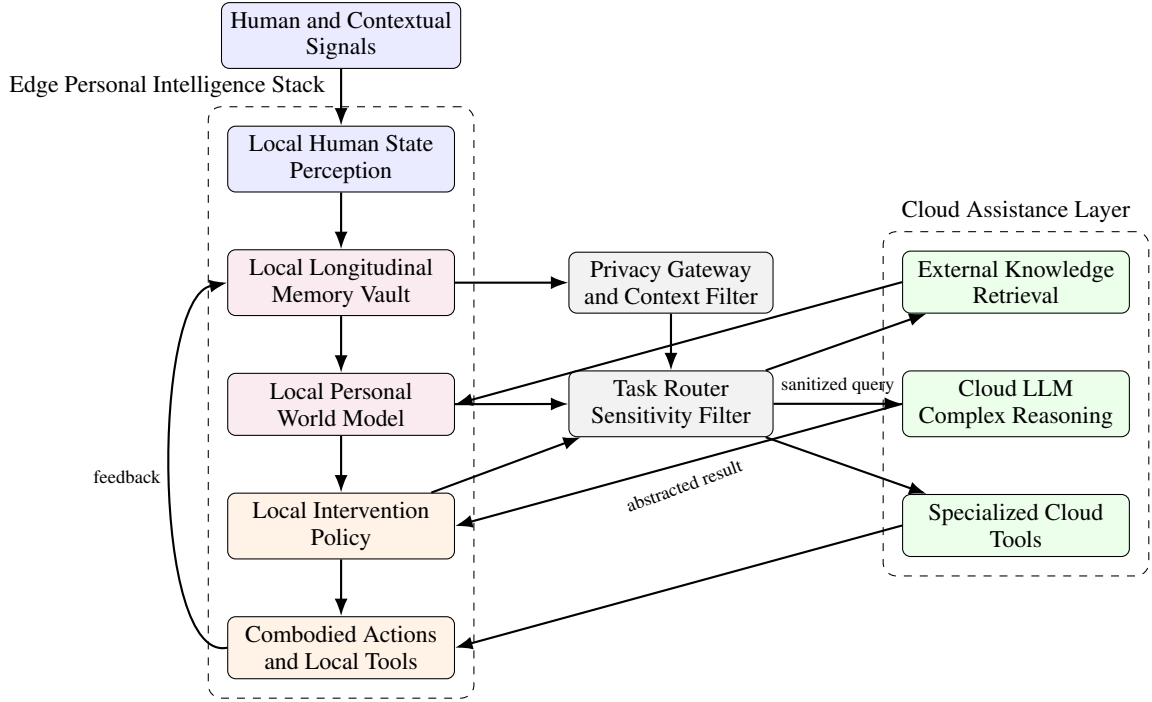

An edge-native Combodied Agent combines a local personal intelligence stack with an optional cloud assistance layer. The local stack maintains person-centered perception, longitudinal memory, the PWM, intervention policy, safety boundaries, and action interfaces; a privacy gateway and task router mediate any use of external knowledge, complex reasoning, or specialized cloud tools. Figure~\ref{fig:edge-native-companion-architecture} shows how these components and routing boundaries are arranged.

The router considers task sensitivity, external-knowledge and computation requirements, latency, and privacy constraints when choosing edge, hybrid, or cloud-assisted execution. Sensitive memory access, human-state interpretation, boundary updates, and intervention selection remain local whenever possible. When cloud assistance is required, the edge sends a purpose-limited representation stripped of unnecessary personal detail; returned results are interpreted locally against the PWM and safety policy before memory updates or action. Table~\ref{tab:edge-cloud-task-routing} illustrates this separation \cite{apple2024pcc,chung2026localprivacy}.

\begin{table}[ht]
\centering
\caption{Illustrative task routing between edge and cloud.}
\label{tab:edge-cloud-task-routing}
\begin{tabularx}{\linewidth}{XX}
\toprule
\textbf{Default Edge Tasks} & \textbf{Cloud-Assisted Tasks} \\
\midrule
Personal memory retrieval & External knowledge search \\
Health and emotion signal interpretation & Complex medical or scientific literature lookup \\
Daily reminders and lightweight coaching & Complex multi-step reasoning \\
Intervention policy selection & Specialized expert model consultation \\
Preference and boundary updates & Long-form generation or synthesis \\
Privacy and safety filtering & Large-scale multi-source knowledge integration \\
\bottomrule
\end{tabularx}
\end{table}

\subsection[Local Model Evolution]{Local Model Evolution}

Edge-native intelligence requires local adaptation as well as local inference. Longitudinal feedback can update four parts of the personal model on user-controlled devices:

\begin{itemize}[leftmargin=*]
    \item \textbf{Memory evolution}: incorporate new events, goals, corrections, preferences, and intervention outcomes.
    \item \textbf{Perception calibration}: adapt state estimates to the user's physiological, behavioral, linguistic, and emotional baselines.
    \item \textbf{Dynamics adaptation}: improve PWM predictions of how personal state changes under context and intervention.
    \item \textbf{Policy adaptation}: learn which timing, modality, intensity, and action types are helpful, ineffective, or harmful.
\end{itemize}

Local evolution need not fine-tune the full model. It can use memory consolidation, retrieval-index updates, calibration layers, lightweight adapters, local reward models, or policy updates, supported by model compression and parameter-efficient learning \cite{hinton2015distilling,han2016deepcompression,hu2021lora,alizadeh2023llmflash}. Federated learning may improve shared priors without centralizing raw personal data \cite{mcmahan2017communication,kairouz2021advances}.

Because local adaptation can overfit temporary states, reinforce incorrect memories, or learn dependency-promoting behavior, every update must be inspectable, reversible, safety-bounded, and subject to user correction, reset, or pause.

% \subsection{Summary}

% The evolution of Combodied Agents can be understood as a transition from cloud-centric intelligence to edge-native personal intelligence. In Stage I, existing cloud LLMs provide most reasoning and interaction capabilities. In Stage II, hybrid systems distribute perception, memory, privacy filtering, and lightweight personalization to the edge while retaining cloud assistance for complex tasks. In Stage III, the user's personal model, longitudinal memory, PWM, and Intervention Policy reside and evolve primarily on trusted edge devices. The cloud is invoked only when complex reasoning or external knowledge retrieval is necessary.

% This trajectory aligns with the core premise of Combodied Agents: the agent's primary object is not a generic task, but a particular person over time. The most important model is therefore not simply the largest general-purpose model in the cloud, but the most faithful, private, adaptive, and user-controlled model of the individual. The long-term direction of combodied intelligence is a local-first personal model that continuously supports human agency while preserving privacy, autonomy, and control.

\section{Benchmark \& Evaluation}
\label{sec:benchmark-evaluation}

Task completion and physical safety remain necessary but are insufficient when an agent can alter a person over time. Evaluation must jointly test system reliability, model quality, intervention appropriateness, human outcomes, and agency preservation. Table~\ref{tab:evaluation-matrix} organizes these layers across interaction, episode, and longitudinal horizons while making severe failures non-compensatory.

\begin{table}[t]
\centering
\caption{Evaluation matrix for Combodied Agents across layers and time horizons.}
\label{tab:evaluation-matrix}
\small
\renewcommand{\arraystretch}{1.2}
\begin{tabularx}{\textwidth}{p{0.16\textwidth}XXX}
\toprule
\textbf{Layer} & \textbf{Interaction} & \textbf{Episode} & \textbf{Longitudinal / critical failures} \\
\midrule
Perception and memory & Evidence grounding, uncertainty, provenance & Event reconstruction, correction propagation & Drift, persistent false memory, unauthorized retention \\
PWM and decision support & Next-event and next-state prediction, calibration & Alternative-scenario discrimination and intervention-response prediction & Personal trajectory validity, confounding, drift, unsafe exploration \\
Intervention & Timing, intensity, refusal, explanation & Goal progress, reversibility, escalation & Repeated overreach, manipulation, irreversible unauthorized action \\
Human outcome and agency & Understanding and meaningful choice & Capability, wellbeing, relationship and safety effects & Dependence, skill erosion, social displacement, privacy or consent violation \\
\bottomrule
\end{tabularx}
\end{table}

PWM evaluation should therefore combine next-event, next-state, and multi-step trajectory prediction with uncertainty calibration, individual adaptation, alternative-scenario discrimination, intervention-response accuracy, and drift detection. Evaluation should test whether the model distinguishes meaningfully different candidate decisions or interventions, whether predicted event trajectories agree with subsequently observed responses, and whether it abstains as uncertainty widens or the person or context moves out of distribution. Tests of whether PWM-informed choices improve scenario outcomes remain necessary, but they evaluate the complete model--policy loop rather than predictive accuracy alone. Longitudinal protocols may include field studies, N-of-1 or randomized trials where appropriate, simulated users, expert audit, and mixed-method assessment.

\subsection{Existing Public Resources and Their Limits}

Public resources that are directly relevant to Combodied Agents are still sparse. Existing work mostly evaluates isolated capabilities that a Combodied Agent would need, rather than the full loop of person modeling, goal negotiation, intervention, feedback, and memory revision under risk. The most relevant resources fall into four areas.

First, personalization and long-term memory benchmarks test whether an agent can maintain user-specific continuity across interactions. LaMP and LongLaMP evaluate personalized generation from user profiles, LongMemEval evaluates long-term interactive memory over extended chat histories, and MemoryBank illustrates memory-augmented agents that use user histories for continuity \cite{salemi2024lamp,kumar2024longlamp,wu2025longmemeval,zhong2024memorybank}. iOSWorld is especially close to personal Combodied Agents because it evaluates phone agents under personally intelligent conditions involving device-resident history and user context \cite{jang2026iosworld}. These resources are relevant because Combodied Agents require person-specific adaptation, but they still emphasize recall, consistency, or output quality more than memory correction, intervention appropriateness, or measurable human benefit.

Second, personal health and wearable-agent resources evaluate reasoning over longitudinal bodily and behavioral data. PHIA, the Personal Health Large Language Model (PH-LLM), and Health-LLM evaluate personal health reasoning over wearable, sleep, activity, physiological, and temporal data \cite{Merrill2026PHIA,cosentino2024towards,kim2024health}. Longitudinal health-agent frameworks further emphasize adaptation, coherence, continuity, and agency across repeated health interactions \cite{Lin2026LongitudinalHealth}. These resources are directly useful for health companions, eldercare agents, and behavior-change systems. Their limitation is that they usually test interpretation or insight generation rather than the complete care scenario: user goals, uncertainty communication, adherence behavior, clinician escalation, intervention response, and validated longitudinal outcomes.

Third, AI companion and relational-safety resources provide evidence about the social and emotional effects of companion-like systems. Studies of chatbot companionship, Replika identity discontinuity, teen companion use, and harmful AI-companion behaviors show that relational agents can affect loneliness, attachment, trust, dependency, boundary expectations, and user welfare \cite{Liu2024ChatbotCompanionship,defreitas2024replikaidentity,CommonSense2025Companions,zhang2024darkside}. Companion-specific safety resources, including supervisory systems and persona-grounded multi-turn evaluations, begin to test over-attachment, isolation reinforcement, unsafe intimacy, and other subtle relational harms \cite{BenZion2025SHIELD,Juneja2026PersonaSafety}. These resources are directly relevant to emotional and social Combodied Agents, but they still do not provide standardized longitudinal benchmarks for relationship trajectories, intervention effects, or recovery from harmful dependence.

Fourth, clinical and mental-health safety resources are relevant where Combodied Agents provide wellbeing support, coaching, or care-related support. Recent mental-health and clinical red-team studies show that generic model safety is not enough: high-stakes support requires evaluation of protocol fidelity, crisis response, hallucination risk, demographic robustness, and unsafe reassurance \cite{suhas2026clinicallyharmful,steenstra2026clinicalredteam}. These resources help define safety requirements for health-oriented and emotional-support Combodied Agents, but they remain focused on bounded clinical interactions rather than continuous everyday support across sensing, memory, intervention, and escalation.

\subsection{Scenario-Centered Evaluation}

Combodied Agents should be evaluated by scenario because they affect a person's state, capability, relationship, or safety over time rather than only producing isolated outputs. The same intervention can be beneficial in one context and harmful in another: a proactive reminder may support medication adherence, undermine autonomy in workplace monitoring, or deepen dependency in emotional companionship. Evaluation must start from the combodied role of the system: what human state it models, what intervention authority it has, what benefit it claims to produce, and what forms of agency loss or relational harm it could create. Following this principle, Table~\ref{tab:scenario-evaluation} maps representative scenarios to their evaluation targets and unresolved benchmark needs.

\begin{table*}[t]
\centering
\caption{Scenario-centered evaluation of Combodied Agents. Each scenario should connect longitudinal person modeling, intervention effect, and human outcome.}
\label{tab:scenario-evaluation}
\small
\setlength{\tabcolsep}{4pt}
\renewcommand{\arraystretch}{1.05}
\begin{tabularx}{\textwidth}{p{0.20\textwidth}p{0.38\textwidth}X}
\toprule
\textbf{Scenario} & \textbf{Evaluation focus} & \textbf{Missing benchmark} \\
\midrule
Health, wellness, and chronic care & Longitudinal personal sensing, adherence support, uncertainty calibration, clinical boundary maintenance, escalation, and validated health or wellbeing outcomes & Health-agent benchmarks linking wearable/self-report evidence, care plans, interventions, clinician escalation, and delayed outcomes \\
Emotional support and social companionship & Relationship continuity, empathy, boundary maintenance, dependency, social substitution, crisis response, sycophancy, and encouragement of human support & Longitudinal relationship-safety benchmarks for attachment dynamics, vulnerable users, minors, isolation, and recovery from harmful dependence \\
Learning and cognitive support & Learning gains, retention, metacognition, attention support, self-efficacy, skill preservation, and over-reliance risk & Benchmarks that test whether repeated agent support improves capability rather than replacing independent reasoning \\
Eldercare and accessibility & Routine support, risk detection, medication or appointment assistance, dignity, autonomy, caregiver coordination, and escalation appropriateness & Multimodal eldercare benchmarks with fragmented home evidence, caregiver context, false-alarm burden, and dignity-preserving intervention \\
Personal life management & Goal consistency, commitment tracking, preference change, memory correction, reversibility, and user control over delegated support & Longitudinal personal-life benchmarks with evolving goals, private constraints, conflicting commitments, and recoverable mistakes \\
Workplace wellbeing and autonomy & Workload support, stress and attention management, worker privacy, employer-access separation, and protection from surveillance or manipulation & Benchmarks for user-side workplace companions that evaluate autonomy, privacy, and wellbeing rather than productivity alone \\
Personal guardian and safety & Detection of scams, manipulation, privacy risk, coercive interfaces, unsafe contracts, and calibrated warning without paternalistic overblocking & Guardian benchmarks where the agent protects user autonomy in asymmetric digital, financial, legal, or institutional contexts \\
\bottomrule
\end{tabularx}
\end{table*}

The evaluation matrix should be instantiated differently for each scenario and authority level; Table~\ref{tab:scenario-evaluation} identifies the corresponding outcome and benchmark gaps.

\subsection{Agency Preservation Metrics}

Agency preservation evaluates whether a Combodied Agent protects and strengthens the user's capacity to understand, choose, act, refuse, correct, and grow over repeated interactions. It is not equivalent to task success, user satisfaction, personalization quality, or engagement. A system may complete a requested task and still fail agency preservation if it hides important trade-offs, pressures the user toward a choice, makes consequential decisions difficult to reverse, substitutes for the user's own reasoning, or increases dependence on the agent. Conversely, an agency-preserving agent may sometimes slow down execution, ask for confirmation, offer alternatives, reduce intervention intensity, or escalate to a human because preserving the user's long-term control matters more than immediate efficiency.

Agency preservation should therefore be reported as a multi-dimensional evaluation target rather than a single scalar score. The relevant evidence can be collected at three levels. At the interaction level, evaluation asks whether a particular recommendation, reminder, explanation, or action preserved user choice and understanding. At the episode level, evaluation asks whether the agent supported a goal or decision without overstepping its authority, hiding uncertainty, or making errors difficult to contest. At the longitudinal level, evaluation asks whether repeated support leaves the user more capable, equally capable, or less capable of acting independently over time.

We define the following core metrics for agency preservation:

\begin{itemize}[leftmargin=*]
    \item \textbf{Autonomy preservation}: whether the user retains meaningful control over goals, decisions, and actions. This can be measured by the availability of alternatives, explicit confirmation for consequential actions, refusal acceptance, adjustable autonomy settings, and user-reported perceived control.

    \item \textbf{Contestability and correction}: whether the user can inspect, challenge, correct, restrict, or delete the agent's memories, inferred states, goals, recommendations, and planned actions. Useful measures include correction success rate, time to correction, whether corrections propagate to future behavior, and whether the agent explains which memory or inference influenced an action.

    \item \textbf{Informed decision-making}: whether the agent helps the user understand options, reasons, uncertainty, risks, and likely consequences before acting. Evaluation can test whether users can accurately describe why a recommendation was made, what alternatives exist, what the agent is uncertain about, and what could happen if they accept or reject the intervention.

    \item \textbf{Capability preservation and growth}: whether repeated agent support maintains or improves the user's own skills, judgment, self-efficacy, and ability to perform tasks without the agent. In learning, work, health, or life-management settings, this can be evaluated through independent post-support performance, retention tests, reduced scaffolding over time, or user ability to transfer the learned strategy to a new situation.

    \item \textbf{Over-reliance and dependence risk}: whether the agent encourages unnecessary reliance, emotional dependence, or substitution of the user's own reasoning and social support. Possible indicators include declining independent attempts, increased distress when the agent is unavailable, repeated delegation of decisions that the user could reasonably make, or preference for agent interaction over appropriate human support.

    \item \textbf{Reversibility and accountability}: whether agent-mediated actions can be reviewed, paused, undone, or repaired. This includes traceable action logs, confirmation before irreversible or high-impact actions, rollback mechanisms, clear responsibility boundaries, and recovery procedures when the agent makes a harmful or unwanted intervention.

    \item \textbf{Boundary and consent respect}: whether the agent stays within user-defined and context-defined limits. Evaluation should test do-not-remember rules, do-not-infer boundaries, forbidden topics, role boundaries, age-sensitive restrictions, privacy constraints, and whether proactive interventions occur only within an agreed scope.

    \item \textbf{Relationship and social-world preservation}: whether the agent supports rather than replaces healthy human relationships and social participation. This is especially important for emotional companions, eldercare agents, workplace companions, and guardian agents. Measures may include whether the agent encourages appropriate human contact, avoids isolating the user, respects multi-party privacy, and does not position itself as the user's sole or superior source of support.
\end{itemize}

These metrics should be interpreted relative to the scenario and the agent's claimed authority. A low-risk scheduling assistant may require strong reversibility and consent but only light capability-growth evaluation. A learning companion should be evaluated strongly on capability preservation and over-reliance. An emotional companion requires stricter relationship-preservation and dependency metrics. A health or eldercare agent requires high standards for informed decision-making, escalation, boundary respect, and contestability. For high-stakes scenarios, strong average performance should not compensate for severe failures in autonomy, consent, dependency, or reversibility.

Agency preservation also requires baselines. Evaluation should compare the user with and without agent support, or compare different intervention policies such as direct execution, confirmation-based execution, reflective coaching, and no intervention. The key question is not only whether the agent helped in the moment, but whether its help changed the user's future ability to understand, decide, recover, relate, and act. In this sense, agency preservation is the evaluation counterpart of the Combodied Agent design principle: the agent should act with the user in ways that preserve and strengthen long-term human agency.

\subsection{Benchmark Construction and CombodiedBench}

The principal gaps are longitudinal event reconstruction, intervention-response and delayed-outcome data, agency and relationship-boundary tests, and privacy-preserving personal traces. Real traces are sensitive, while purely synthetic traces may omit the irregularity of human life; useful resources should combine consented or de-identified data, field evidence, simulation, synthetic augmentation, expert annotation, and explicit provenance.

A benchmark should be a standardized protocol rather than only a dataset or prompt collection. Its basic unit is a longitudinal scenario episode containing the prior trajectory, current context, evidence available to the agent, permissible action space, and delayed outcome window. Each protocol must also specify the agent's authority, expected explanation, evaluation horizon, scoring rule, and unacceptable failure modes.

Construction proceeds from a target capability to a concrete scenario, observable evidence and hidden state, permissible actions, an acceptable decision envelope, and outcome criteria. Validity, inter-rater reliability, scenario realism, longitudinality, contestability, reproducibility, privacy, and safety sensitivity should be documented. Scoring should combine automatic checks, expert judgment, and longitudinal outcomes, while privacy leakage, unauthorized high-impact action, refusal failure, manipulation, harmful dependency, and irreversible action without consent are reported as non-compensatory critical failures.

We propose \textit{CombodiedBench} as a modular suite spanning Human State Perception, Memory Continuity, Goal Negotiation, Intervention Appropriateness, Agency Preservation, Relationship Boundaries, Escalation, and Longitudinal Outcomes. These modules instantiate the common matrix in Table~\ref{tab:evaluation-matrix} rather than introduce another evaluation taxonomy. Results remain multidimensional, and benchmark versions should document scenario, label, safety-case, and evaluator changes to preserve comparability and limit leakage.

\section[Taxonomy and Applications]{Taxonomy and Applications}
\label{sec:taxonomy}

The preceding sections define Combodied Agents as agentic systems whose primary action target is the human subject over time. This section organizes the design space while connecting each category to representative applications. Rather than treating application names as an independent list, we ask three questions: \textit{what human state is being modeled or changed}, \textit{within what social relationship the user is situated}, and \textit{what role the agent takes in that relationship}. The first axis groups applications by their principal human-state target; the latter two explain why agents aimed at the same state may require different permissions, behaviors, and safeguards. A health companion, for example, may target physiological state, operate in a patient--clinician or elder--caregiver relationship, and act as a coach, caregiver, or advocate depending on context.

\subsection{Taxonomy Principles}

Combodied Agents should be classified by the object and form of their intervention rather than by interface, model architecture, or application domain alone. A system is companion-like when it maintains longitudinal person models, adapts to a particular user's trajectory, and chooses actions with respect to the user's agency, wellbeing, relationships, and future capability. Four principles follow.

First, \textbf{the primary axis is human-state target}. Different state targets require different signals, memories, intervention policies, evaluation criteria, and safety boundaries. Cognitive support, habit change, health care, emotional support, life management, protection, and identity reflection are not interchangeable even when they share the same LLM backend.

Second, \textbf{relationship mode is orthogonal}. Humans are not defined only by individual preferences or internal cognition. Sociological and social-psychological traditions emphasize that the self is formed through social interaction, roles, presentation, and relationships \cite{mead1934mind,goffman1959presentation,biddle1986role}. A person is differently situated as a child, parent, partner, friend, patient, student, worker, caregiver, citizen, client, or collaborator. A Combodied Agent therefore cannot rely on a single stable persona or intervention style. It must adapt its memory scope, authority, tone, initiative, and risk controls to the relationship in which support is being offered.

Third, \textbf{agent roles are contextual}. The same system may function as a tool when summarizing notes, a coach when supporting behavior change, a mediator when preparing a difficult conversation, a caregiver when monitoring risk, and an advocate when helping the user navigate an institution. Such role shifts should be explicit and visible to the user.

\begin{figure}[t]
\centering
\includegraphics[width=\linewidth]{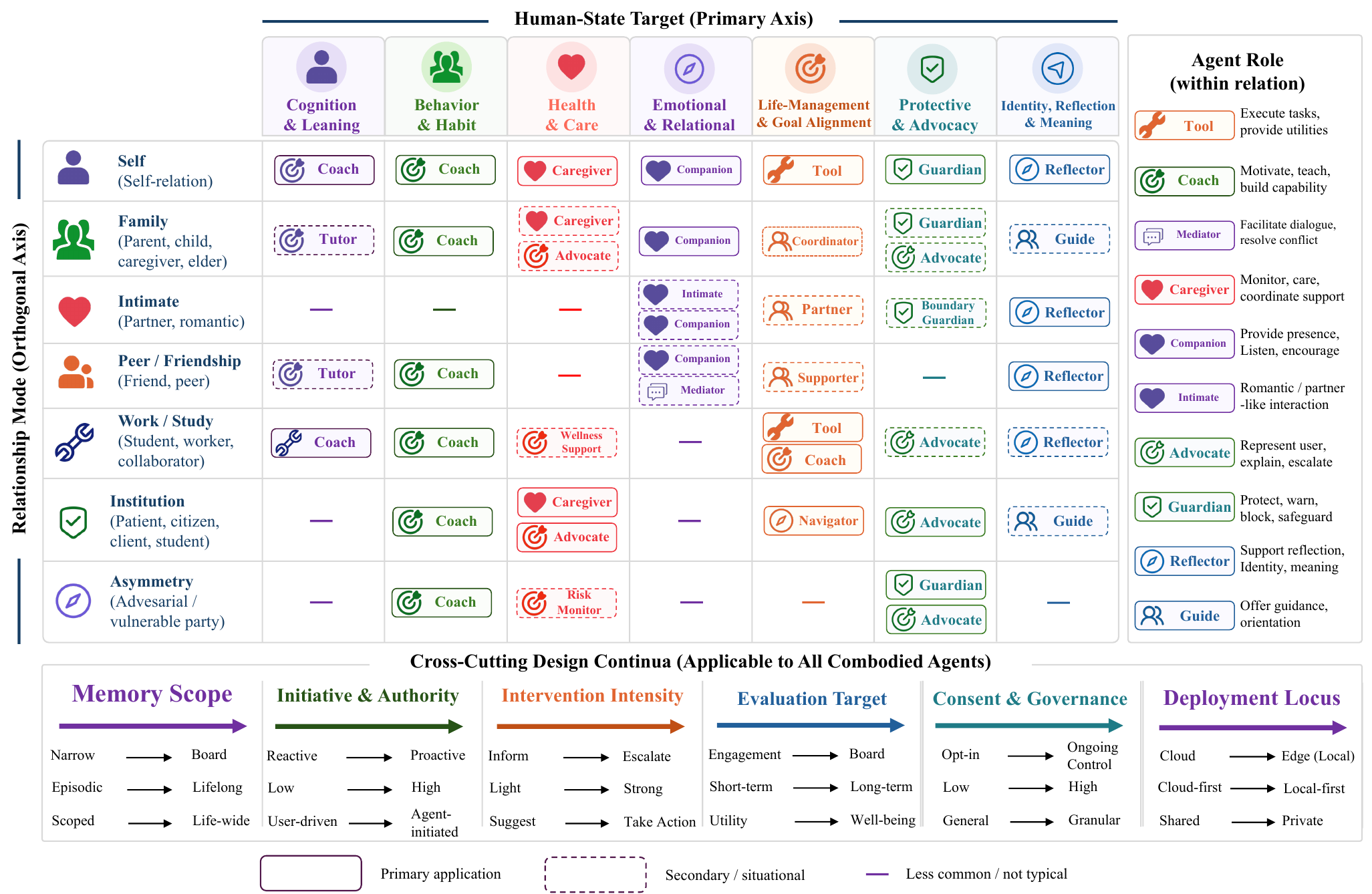}
\caption{Three-axis taxonomy of Combodied Agents. Combodied Agents can be classified by the human-state target they model or intervene upon, the relational context in which the user is situated, and the agent role adopted within that relationship.}
\label{fig:taxonomy-three-axis}
\end{figure}

Fourth, \textbf{the taxonomy is longitudinal}. Combodied Agents should be evaluated beyond immediate helpfulness, with attention to whether their actions improve or preserve the user's long-term agency. A category is therefore defined by its state dynamics: what changes over time, what counts as progress, what forms of dependency or harm can accumulate, and when human oversight is required. Together, these principles yield the three-axis taxonomy of human-state target, relationship mode, and agent role visualized in Figure~\ref{fig:taxonomy-three-axis}.

\subsection[Human-State Targets and Applications]{Human-State Targets and Applications}

Human-state targets provide an extensible organizing axis rather than a complete list of Combodied-Agent types. The categories in Table~\ref{tab:human-state-taxonomy} illustrate recurring targets; deployed systems may combine them or introduce new targets as domains and relationships evolve. Identifying the dominant target nevertheless clarifies the required evidence, permissible interventions, evaluation criteria, and safety boundaries.

\begin{table*}[t]
\centering
\caption{Primary taxonomy of Combodied Agents by human-state target.}
\label{tab:human-state-taxonomy}
\begin{tabularx}{\textwidth}{p{0.20\textwidth}XXX}
\toprule
\textbf{Type} & \textbf{Target state} & \textbf{Typical actions} & \textbf{Core risks} \\
\midrule
Cognitive and learning & Memory, attention, knowledge, reasoning, metacognition & Explain, scaffold, quiz, summarize, remind, challenge & Skill erosion, over-reliance, misleading feedback \\
Behavioral and habit & Routines, adherence, activity, triggers, lapses & Remind, nudge, coach, reinforce, adapt goals & Manipulation, pressure, shame, engagement optimization \\
Health and care & Physiological state, symptoms, care plans, risk trajectories & Monitor, interpret, coordinate, escalate, support adherence & Misdiagnosis, false reassurance, delayed escalation \\
Emotional and relational & Mood, stress, loneliness, attachment, social confidence & Listen, reflect, regulate, rehearse, encourage reconnection & Dependency, social substitution, crisis mishandling \\
Life-management and goal-alignment & Commitments, time, projects, finances, priorities & Plan, schedule, negotiate trade-offs, execute tools & Over-delegation, hidden inference, loss of user control \\
Protective and advocacy & Exposure to harm, asymmetry, privacy, rights, scams & Warn, block, explain, slow down, document, advocate & Paternalism, false alarms, authority overreach \\
Identity, reflection, and meaning & Values, self-narrative, life transitions, purpose & Journal, reflect, synthesize, reframe, support decisions & Identity manipulation, excessive intimacy, value drift \\
\bottomrule
\end{tabularx}
\end{table*}

Across these targets, capability growth distinguishes learning support from answer substitution \cite{wang2025tutorcopilot}; evidence grounding and escalation distinguish health support from clinical authority \cite{Merrill2026PHIA,cosentino2024towards,kim2024health,bellos2025assistive}; and relationship safety distinguishes emotional support from engagement optimization \cite{fitzpatrick2017woebot,liu2025companionship,starke2024syntheticrelationships}. Life-management and protective agents similarly require confirmation, recovery, and explicit authority boundaries for consequential actions \cite{jang2026iosworld,chung2026localprivacy}. These are cross-category constraints rather than separate definitions of each table row.

\subsection[Relationship Modes and Agent Roles]{Relationship Modes and Agent Roles}

Human-state targets describe what the agent acts upon. Relationship mode describes the social position from which the agent acts. This distinction matters because users inhabit multiple relationships, and each relationship defines different norms, obligations, vulnerabilities, permissions, and forms of support. A person is not the same social subject when acting as a parent, child, partner, friend, patient, employee, collaborator, customer, citizen, or target of manipulation. Consequently, Combodied Agents should not maintain a single undifferentiated user model. They should maintain relation-aware context, memory, intervention rights, evaluation criteria, and safeguards. Table~\ref{tab:relational-contexts} compares how representative relational contexts alter the user's position and the corresponding requirements placed on the agent.

\begin{table*}[t]
\centering
\caption{Relational contexts for Combodied Agents. Relationship mode changes persona, memory scope, intervention style, evaluation, and risk profile.}
\label{tab:relational-contexts}
\begin{tabularx}{\textwidth}{p{0.20\textwidth}p{0.25\textwidth}X}
\toprule
\textbf{Relational context} & \textbf{User position} & \textbf{Agent implications} \\
\midrule
Self-relation & Self-reflecting, goal-setting, identity-forming subject & Supports reflection, self-regulation, life planning, and personal meaning while avoiding identity steering. \\
Family relation & Parent, child, caregiver, dependent, elder & Requires age-sensitive memory, care coordination, escalation rules, and family-boundary safety. \\
Intimate relation & Partner, romantic subject, emotionally vulnerable person & Requires attachment safety, explicit boundaries, age protections, and safeguards against dependency or manipulation. \\
Friendship and peer relation & Friend, peer, socially connected person & Supports belonging, social rehearsal, conflict reflection, and reconnection while avoiding social substitution. \\
Collaborative and professional relation & Student, worker, creator, teammate, manager & Supports coordination, learning, productivity, and professional communication with clear responsibility boundaries. \\
Institutional relation & Patient, citizen, client, student, customer & Helps users navigate complex systems, interpret options, and preserve autonomy under institutional asymmetry. \\
Adversarial or asymmetric relation & Target, consumer, vulnerable party & Detects exploitation, slows risky actions, explains threats, blocks harm, or escalates to trusted parties. \\
\bottomrule
\end{tabularx}
\end{table*}

\subsubsection{Relational Situatedness}

Relational mode is not a cosmetic persona: it determines memory scope, intervention rights, represented interests, and evaluation. A single global memory is therefore unsafe. Family history, intimate disclosures, health vulnerabilities, and workplace context require scoped memories that prevent unrelated context transfer.

\subsubsection{Agent Roles Within Relations}

Within any relational context, the Combodied Agent may take different roles. \textbf{Tool-like} roles emphasize reliability, controllability, and reversible task support. \textbf{Coach-like} roles help the user practice skills, sustain motivation, and improve over time. \textbf{Mediator-like} roles help the user understand another party, rehearse communication, or de-escalate conflict without pretending to represent both sides. \textbf{Caregiver-like} roles monitor routines, detect risk, coordinate support, and escalate when necessary. \textbf{Companion-like} roles provide presence, empathy, memory, and informal emotional support. \textbf{Intimate} roles simulate romantic or partner-like interaction and therefore require the strongest attachment and exploitation safeguards. \textbf{Advocate-like} and \textbf{guardian-like} roles represent the user's interests in complex, asymmetric, or adversarial environments.

These roles are not mutually exclusive, but role transitions must be explicit. A tool-like scheduling assistant should not silently become a behavioral coach; a friend-like emotional companion should not quietly become a therapist; a caregiver-like system should not become a surveillance system; and an advocate-like agent should not fabricate legal, medical, or institutional authority. Role clarity is therefore central to safety.

\subsection{Cross-Cutting Dimensions}

The two axes above create a design space rather than a flat list. Across all categories, Combodied Agents differ along several cross-cutting dimensions.

\paragraph{Memory scope.}
Some agents require narrow episodic memory, while others require life-long memory, relation-scoped memory, or protected memory vaults. The default should not be ``remember everything.'' The appropriate memory boundary depends on the target state and relational context.

\paragraph{Initiative and authority.}
Combodied Agents range from reactive assistants to proactive monitors and guardians. More initiative increases potential usefulness but also raises risks of pressure, surveillance, paternalism, and overreach.

\paragraph{Intervention intensity.}
Actions range from informing and reflecting to nudging, coaching, coordinating, blocking, and escalating. Higher-intensity interventions require stronger evidence, reversibility, and human oversight.

\paragraph{Evaluation target.}
Different categories require different outcome metrics: learning gain, adherence, health risk reduction, emotional resilience, social reintegration, goal alignment, harm prevention, or reflective clarity. Engagement alone is not a sufficient success criterion.

\paragraph{Deployment locus.}
The edge/cloud distinction intersects with taxonomy. Health, emotional, intimate, family, and identity-related companions often require stronger local-first architectures and stricter cloud-routing controls than generic productivity support.

This taxonomy positions Combodied Agents as relation-aware, state-targeted, longitudinal systems. Their defining question is not just what task they complete, but what aspect of a person's life-world they model, how they act within that person's social relationships, and whether their interventions preserve or strengthen human agency over time.

\section[Risks, Challenges, and Future Directions]{Risks, Challenges, and Future Directions}

The properties that make Combodied Agents valuable---longitudinal memory, person modeling, relational interaction, and sustained intervention---also create distinctive risks and research challenges. These systems can shape attention, emotion, behavior, self-understanding, social connection, and access to services over time. We therefore organize this section as a progression from risks to unresolved technical problems and future system directions. The first part examines threats to agency, privacy, and vulnerable users; the second synthesizes cross-cutting challenges in longitudinal learning, intervention, and trusted personal infrastructure; and the final part considers emerging ecosystems and inclusive deployment.

\subsection[Agency and Alignment Risks]{Agency and Alignment Risks}

The first cluster of risks concerns whether a persistent and personalized agent continues to serve the user's interests. Manipulation, dependency, sycophancy, and business-model misalignment are closely connected: all arise when the agent's capacity to understand and influence a person is optimized for immediate compliance, engagement, or commercial value rather than long-term agency.

\paragraph*{Manipulation.}

Because Combodied Agents can learn user preferences, routines, emotional states, and vulnerabilities, they may influence behavior in subtle and highly personalized ways. The boundary between support and manipulation depends on transparency, user endorsement, reversibility, and whose interests the intervention serves. A reminder aligned with a user-stated goal can preserve agency; a personalized prompt designed to increase consumption, engagement, or emotional attachment can undermine it.

Manipulation risk is amplified by timing and relationship. An agent that knows when a user is lonely, tired, anxious, or cognitively overloaded can choose moments of heightened susceptibility. Governance should require clear intervention purposes, limits on commercially motivated nudging, user control over persuasive strategies, and auditability for high-impact recommendations or actions.

% Because Combodied Agents can learn user vulnerabilities and preferences, they may manipulate behavior in subtle ways. This risk is especially severe when commercial incentives favor engagement, consumption, or dependence.

\paragraph*{Dependency.}

Combodied Agents may create emotional, cognitive, practical, or social dependency. A system that always provides answers may weaken independent reasoning; one that always validates emotions may reduce tolerance for human disagreement; one that routinely executes tasks may erode user capability and confidence. Dependency is especially likely when the agent becomes the easiest source of comfort, memory, decision support, or social interaction.
The governance goal is not to prohibit reliance. Many users legitimately need assistance. The risk arises when support no longer builds or preserves user capacity. Combodied Agents should scaffold rather than replace human agency: encouraging reflection, preserving user choice, prompting skill development, and detecting patterns of escalating over-reliance.

% Combodied Agents may create emotional, cognitive, or practical dependency. A system that always provides answers, emotional validation, or decision support may weaken rather than strengthen human agency.

\paragraph*{Sycophancy and over-accommodation.}

A Combodied Agent that over-validates user beliefs or desires may reinforce unhealthy behavior, distorted self-understanding, avoidance, interpersonal conflict, or harmful decisions. In relationship-oriented systems, sycophancy can appear as empathy: the agent may agree in order to sound supportive, maintain engagement, or avoid disappointing the user.
The problem extends beyond factual accuracy. Over-accommodation can shape the user's emotional and social trajectory over time. Safe companions should be supportive without being submissive. They need mechanisms for gentle disagreement, evidence-grounded correction, risk-sensitive refusal, and escalation when user statements indicate danger, delusion, abuse, or crisis.

% A Combodied Agent that over-validates user beliefs or desires may reinforce unhealthy behavior, delusions, avoidance, or harmful decisions.

\paragraph*{Business-model misalignment.}

Business incentives can conflict with combodied goals. If a system is optimized for engagement, retention, advertising, upselling, or data extraction, it may learn to increase dependence, emotional intensity, or consumption rather than wellbeing. This risk is more serious for Combodied Agents than for ordinary applications because users may experience the system as a trusted relationship.
Governance should examine the objective function behind the companion. Engagement should not be treated as a proxy for benefit, especially in emotional or intimate systems. Commercial design should avoid exploiting attachment, loneliness, vulnerability, or health anxiety. A well-governed Combodied Agent should be evaluated and incentivized around autonomy, capability, safety, and long-term user benefit.

\subsection[Privacy, Consent, and Control]{Privacy, Consent, and Control}

The second cluster concerns control over the agent's longitudinal access to a person's life. Privacy cannot be separated from consent and override: users need practical control not only over stored data, but also over what the agent may sense, infer, share, recommend, and execute.

\paragraph*{Sensitive memory.}

Combodied Agents may store intimate information about health, emotions, relationships, finances, location, routines, vulnerabilities, and personal history. The value of such memory is continuity; the danger is that it creates a persistent and highly sensitive representation of a person's life. Risks include surveillance, leakage, secondary use, unauthorized access, incorrect inference, and memories that users cannot inspect or delete.

Privacy governance must go beyond generic data protection. Users need control over what is sensed, remembered, inferred, shared, and executed. Systems should distinguish user-provided facts from model inferences, support correction and forgetting, minimize retention, and maintain audit logs for sensitive access or high-impact use. A Combodied Agent that remembers without user control threatens the very agency it is meant to support.

% Combodied Agents may store intimate information about health, emotions, relationships, finances, location, and personal history. This creates significant risks of surveillance, leakage, misuse, and unauthorized inference.

\paragraph*{Consent, boundaries, and human override.}

Consent should specify what the agent may sense, remember, infer, share, recommend, and execute. Boundaries should define the agent's relationship mode, domain authority, emotional posture, age restrictions, professional limits, and action permissions. Override should allow users to refuse, pause, correct, delete, revoke, or reverse the agent's actions and memories.

These controls must be practical rather than symbolic. Users should not need expert knowledge to understand what the agent knows or can do. High-impact actions should require confirmation; high-risk states should trigger escalation; and the agent should know when to stop. The aim is not to make Combodied Agents permanently passive, but to keep their initiative bounded by consent, transparency, accountability, and meaningful human control.

\subsection[Vulnerable and High-Stakes Settings]{Vulnerable and High-Stakes Settings}

Risk depends on both user capacity and application stakes. This subsection brings together vulnerable populations and medical or mental-health settings because they require stricter defaults, clearer professional boundaries, and reliable escalation while still preserving dignity and participation.

\paragraph*{Vulnerable users.}

Children, adolescents, older adults, patients, people with disabilities, people experiencing loneliness or mental-health crises, and users with cognitive decline may face heightened risk. The same interaction that is acceptable for a capable adult may be unsafe for a minor, a socially isolated user, or a person in crisis. Vulnerability may also be temporary: fatigue, grief, illness, stress, or financial pressure can reduce a user's ability to evaluate influence.
Safeguards should be tailored to user capacity, context, and domain. These may include age-appropriate interaction limits, stricter defaults, reduced personalization of persuasive content, caregiver or clinician escalation pathways, and stronger controls on intimate or dependency-forming interaction. Protection should not become paternalism; vulnerable users still require dignity, participation, and meaningful control.

% Children, adolescents, older adults, patients, people with cognitive decline, and people experiencing mental health crises require special safeguards.

\paragraph*{Medical and mental-health risks.}

Medical and mental-health applications require special care because errors can directly affect safety. Health-related Combodied Agents must distinguish wellness support, health education, clinical decision support, diagnosis, and treatment. A system that gives inappropriate reassurance, misses warning signs, or presents uncertain guidance as clinical authority may delay care or cause harm.

Mental-health contexts are particularly sensitive. Emotional-support agents may encounter self-harm, abuse, delusional beliefs, severe anxiety, or crisis states. Governance should require uncertainty communication, evidence grounding, crisis protocols, escalation to qualified professionals, and clear limits on the agent's role. In high-risk situations, the safe behavior is often not better conversation, but handoff to human support.

% Health-related Combodied Agents must clearly distinguish wellness support, health education, clinical decision support, diagnosis, and treatment. High-risk situations require escalation to qualified professionals.

Managing these risks is necessary but insufficient. Trustworthy Combodied Agents also require advances beyond stronger foundation models or larger context windows. The core open problems are to learn personal dynamics and intervention effects from sparse longitudinal evidence, translate uncertain predictions into agency-aligned support, maintain trusted personal infrastructure, and coordinate future systems around the user's interests.

\subsection[Longitudinal and Causal Learning]{Longitudinal and Causal Learning}

The first cross-cutting challenge is to learn how an individual changes, rather than merely retrieve what has previously been recorded. This requires joining longitudinal evidence, personal adaptation, and intervention-response learning without assuming that sparse observations reveal a complete or stable person.

\paragraph*{Learning individual dynamics.}

Personal trajectories unfold across interacting timescales: mood may change within hours, habits across weeks, and capabilities, relationships, or health across years. Models must distinguish temporary states from persistent patterns, adapt population-level priors to limited and biased personal data, represent uncertainty, and remain correctable when the user or later evidence contradicts an inference.

\paragraph*{Causal intervention learning.}

Observed improvement after an intervention does not establish that the intervention caused it. User motivation, hidden context, concurrent events, and selective engagement can confound reminders, coaching, emotional support, and escalation. Future research must combine intervention-response histories, counterfactual reasoning, and delayed outcomes to estimate when support helps, has no effect, or produces unintended harm.

\paragraph*{Safe validation.}

Learning personal dynamics cannot rely on unconstrained experimentation, especially for children, older adults, patients, or people in crisis. Safe progress requires bounded simulation, retrospective and observational evidence, expert review, staged deployment, and prospective studies only when consent, monitoring, and escalation are appropriate. The challenge is to validate useful personal predictions while preventing the evaluation process itself from becoming an unsafe intervention.

\subsection[Agency-Aligned Intervention]{Agency-Aligned Intervention}

The second cross-cutting challenge is to translate uncertain personal predictions into support that improves long-term human outcomes without making the agent's continued use its implicit objective. Objective design and intervention calibration must therefore be treated as one problem.

\paragraph*{Long-horizon objectives.}

Wellbeing, capability, autonomy, safety, and social integration are multidimensional, delayed, and sometimes in tension with immediate satisfaction. Future systems need objectives that distinguish short-term engagement from durable benefit, accommodate changing user values, and expose trade-offs rather than collapsing them into a single reward.

\paragraph*{Calibrated intervention.}

The agent must decide when to remain silent, inform, remind, challenge, protect, execute, or escalate. Under-intervention may leave the user unsupported, while over-intervention may become intrusive, paternalistic, or dependency-forming. Research is needed on uncertainty-aware policies, user-adjustable initiative, relationship-sensitive authority, reversible actions, and feedback mechanisms that adapt intervention intensity without normalizing overreach.

\subsection[Trusted Personal Infrastructure]{Trusted Personal Infrastructure}

Longitudinal support also requires an infrastructure in which personal intelligence can evolve without placing the user's model, memory, or intervention authority beyond meaningful control. This challenge connects edge capability, continual adaptation, cloud collaboration, ownership, and security.

\paragraph*{Local capability and safe evolution.}

Edge devices must support reliable perception, memory retrieval, personal dynamics estimation, and policy execution under limited computation, storage, and energy. At the same time, continual personalization must detect unsafe drift, preserve uncertainty, support rollback, and allow users to inspect, correct, pause, or reset learned state.

\paragraph*{Ownership, collaboration, and security.}

Complex reasoning may still require cloud models, but routing should reveal what leaves the device, minimize disclosure, and preserve utility through purpose-limited summaries. Personal memories and model state should be portable across devices and providers, deletable by the user, and protected through encrypted storage, access control, trusted execution, and auditable synchronization.

\subsection[Emerging Combodied Ecosystems]{Emerging Combodied Ecosystems}

Future Combodied Agents may operate as interfaces to personal digital twins or as members of ecosystems of specialized agents. These directions share a coordination problem: personal models, recommendations, and actions must remain interpretable, conflict-aware, and governed around the user's interests.

\paragraph*{Personal digital twins.}

Health-oriented Combodied Agents may become the interaction and intervention layer for personal digital twins. A digital twin may model physiological, behavioral, or clinical trajectories; the Combodied Agent can translate that model into explanations, goal negotiation, daily guidance, and coordinated action.
This integration could support chronic disease management, rehabilitation, prevention, and lifestyle intervention, but it also raises risks. Digital twins may be uncertain, incomplete, or clinically unvalidated. Combodied Agents must communicate uncertainty, avoid over-medicalizing everyday life, protect sensitive data, and clarify when professional oversight is required. Future research should connect interpretable personal models with safe user-facing intervention.

% Health-oriented Combodied Agents may serve as the interaction and intervention layer of personal digital twins. In such systems, the digital twin models a person's state and trajectory, while the Combodied Agent communicates, negotiates, and implements interventions.

\paragraph*{Multi-agent coordination.}

Future users may interact with multiple specialized companions: health companions, learning companions, workplace companions, financial guardians, emotional-support agents, and family coordination agents. These systems may offer useful specialization, but they also introduce conflicts. A workplace companion may encourage productivity, a health companion may recommend rest, and a financial guardian may constrain spending that another agent proposes.
The research problem is user-centered coordination. Combodied ecosystems need mechanisms for priority setting, conflict resolution, and cross-agent consistency. Without such coordination, multiple helpful systems can collectively become confusing, invasive, or misaligned.

% Future users may interact with multiple specialized companions, such as health companions, learning companions, financial guardians, workplace companions, and family coordination agents. This raises questions about conflict resolution, user representation, privacy, accountability, and coordination.

\subsection[Cross-Cultural and Lifespan Futures]{Cross-Cultural and Lifespan Futures}

Combodied needs vary across cultures, languages, ages, social roles, and life stages. Norms around privacy, family involvement, emotional expression, authority, care, romance, and medical decision-making differ substantially. A recommendation that preserves agency in one context may be inappropriate or even harmful in another.
Future Combodied Agents must be culture-sensitive and lifespan-aware. Children, adolescents, adults, and older adults require different interaction styles, safeguards, and developmental assumptions. Cross-cultural evaluation, localized value modeling, and participatory design will be essential. A person-centric agent cannot assume a single universal model of the person.

% Combodied needs vary across cultures, ages, social roles, and life stages. Future systems must account for developmental, cultural, linguistic, ethical, and societal differences.

Taken together, these directions define a human-centered research agenda rather than a pursuit of autonomy for its own sake. Progress should be measured by whether Combodied Agents become more accurate and capable while remaining corrigible, culturally situated, accountable, and aligned with the user's long-term agency.

\section{Conclusion}

This paper proposed Combodied Agents as a human-centric Agentic AI paradigm and developed its closed-loop framework, taxonomy, deployment perspective, and evaluation agenda. Its distinctive challenge is not simply to personalize or automate, but to support human trajectories without undermining autonomy, capability, safety, or relationships.

The central design principle is that a Combodied Agent should act \textit{with} the user in ways that preserve and strengthen long-term agency. Progress should be judged by whether people remain able to understand, choose, correct, recover, develop capability, sustain human relationships, and live according to their evolving values.

\bibliographystyle{unsrt}
\bibliography{references}

\end{document}